\documentclass[3p,twocolumn]{elsarticle}

\makeatletter
\def\ps@pprintTitle{%
 \let\@oddhead\@empty
 \let\@evenhead\@empty
 \def\@oddfoot{\centerline{\thepage}}%
 \let\@evenfoot\@oddfoot}
\makeatother

\usepackage{filecontents}
\RequirePackage{fix-cm}
\usepackage[T1]{fontenc}
\usepackage[utf8]{inputenc}
\usepackage{textcomp}
\usepackage{amsmath}
\usepackage{mathrsfs}
\usepackage [english]{babel}
\usepackage [autostyle, english = american]{csquotes}
\usepackage{graphicx}
\usepackage{multicol}
\usepackage[table]{xcolor}
\usepackage{collcell}
\usepackage{array, hhline}
\usepackage{pgf}
\usepackage{float}
\usepackage[ruled,linesnumbered]{algorithm2e}
\usepackage{subcaption}
\usepackage{amssymb}
\usepackage{xspace}
\usepackage{xtab,afterpage}
\usepackage{multirow}
\usepackage{hyperref}
\graphicspath{{Figures/}}
\newcommand*{\colorme}[1]{%
    \pgfmathparse{#1<.5?1:0}%
    \ifnum\pgfmathresult=0\relax\color{white}\fi
    \pgfmathparse{1-#1}
    \expandafter\cellcolor\expandafter[\expandafter\gray\expandafter]\expandafter{\pgfmathresult}%
    #1%
}

\newcommand{\latinphrase}[1]{\textit{#1}} 

\newcommand{\ie}{\latinphrase{i.e.}\xspace}

\newcommand{\eg}{\latinphrase{e.g.}\xspace}
\newcommand{\etc}{\latinphrase{etc.}\xspace}

\newcommand*{\gray}{gray}

\begin{document}

\title{Survey: Image Mixing and Deleting for Data Augmentation}
\author[1]{Humza Naveed\fnref{fn1}}
\ead{humza\_naveed@yahoo.com}
\author[2]{Saeed Anwar}
\ead{saeed.anwar@kfupm.edu.sa}
\author[3]{Munawar Hayat}
\ead{munawar.Hayat@monash.edu.au}
\author[1]{Kashif Javed}
\ead{kashif.javed@uet.edu.pk}
\author[4]{Ajmal Mian}
\ead{ajmal.mian@uwa.edu.au}

\address[1]{University of Engineering and Technology, Lahore, Pakistan}
\address[2]{King Fahad University of Petroleum and Minerals, Dhahran 31261, KSA}
\address[3]{Monash University, Melbourne, VIC 3800, Australia}
\address[4]{University of Western Australia, Perth, Australia}

\fntext[fn1]{Corresponding author.}

\begin{abstract}
Neural networks are prone to overfitting and memorizing data patterns. To avoid over-fitting and enhance their generalization and performance, various methods have been suggested in the literature, including dropout, regularization, label smoothing, \etc One such method is augmentation which introduces different types of corruption in the data to prevent the model from overfitting and to memorize patterns present in the data. A sub-area of data augmentation is \emph{image mixing and deleting}. This specific type of augmentation either deletes image regions or mixes two images to hide or make particular characteristics of images confusing for the network, forcing it to emphasize the overall structure of the object in an image. Models trained with this approach have proven to perform and generalize well compared to those trained without image mixing or deleting. An added benefit that comes with this method of training is robustness against image corruption. Due to its low computational cost and recent success, researchers have proposed many image mixing and deleting techniques. We furnish an in-depth survey of image mixing and deleting techniques and provide categorization via their most distinguishing features. We initiate our discussion with some fundamental relevant concepts. Next, we present essentials, such as each category's strengths and limitations, describing their working mechanism, basic formulations, and applications. We also discuss the general challenges and recommend possible future research directions for image mixing and deleting data augmentation techniques.  Datasets and codes for evaluation are publicly available \href{https://github.com/humza909/Survery-Image-Mixing-and-Deleting-for-Data-Augmentation}{here}.

\begin{keyword}
Image Augmentation \sep Data Augmentation \sep Image Mixing \sep Data Deleting \sep Survey \sep Review
\end{keyword}
\end{abstract}

\maketitle
  
\begin{figure*}[!tbp]
\centering
\includegraphics[width=1.0\textwidth]{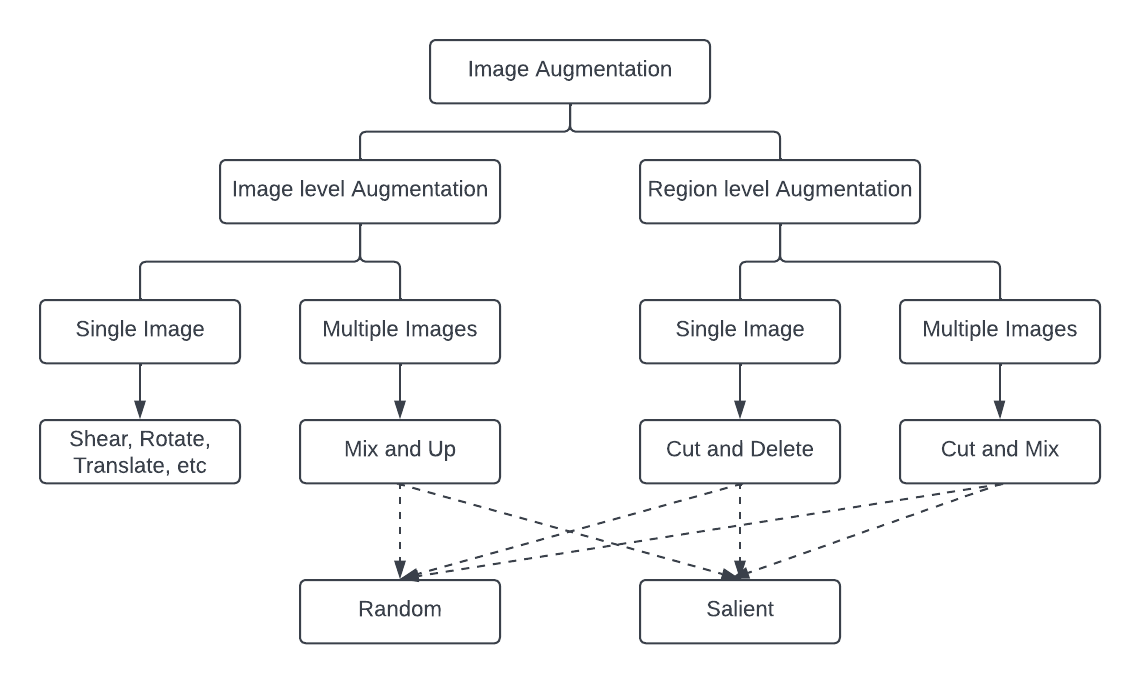}
\caption{The categorization of image mixing and deleting data augmentation techniques. The three main types are Cut $\&$ Delete, Cut $\&$ Mix and Mix $\&$ Up.}
\label{flow_diagram}
\end{figure*}

\section{Introduction}
\begin{flushright}
\enquote{Where there is ruin, there is hope for a treasure.} - Rumi
\end{flushright}

Deep Neural Networks have shown tremendous success in image classification~\cite{imagenet,clf}, object detection~\cite{fasterrcnn,yolov3}, and semantic segmentation~\cite{deeplab,semanticseg}. These
networks are data hungry with millions of parameters that make them prone to overfitting~\cite{overfitting2,overfitting3,overfitting}.
To enhance model generalization, many approaches
have been suggested, for example, regularization~\cite{regularization},
dropout~\cite{dropout}, and data augmentation~\cite{surveyimaug}. The purpose of data augmentation is to increase the dataset size by introducing various corruptions in the data so that the model does not memorize the irrelevant patterns in the data. Conventional image augmentations include image rotation, random cropping, jittering, translation, shear, solarize, posterize, brightness, \etc
In addition to these techniques, a relatively new data
augmentation strategy mixes and deletes different image regions. The
augmented images generated by this approach are present in the vicinity of the original data distribution. The model trained with this new data generalizes well against data variations, avoids overfitting, and
achieves robustness against data corruption. The benefits of image mixing and deleting are label smoothing, robustness to occlusions, noise, image corruptions, and adversarial examples, improved classifier calibration, consistency, and out-of-distribution data predictions. Overall, the model learns to focus on the complete structure of the object under consideration resulting in improved performance.

Image mixing and deleting can be divided into three main categories: 1) cut and delete 2) cut and mix, and
3) mix and up. Figure~\ref{flow_diagram} shows the taxonomy of image augmentation. Cut and delete~\cite{cutout,hidenseek,gridmask,rerase} erases
the selected region, cut and mix~\cite{cutmix,attentivecutmix,ricap,resizemix,saliencymix,snapmix} 
replaces the selected area with some part of another image, whereas mix and up~\cite{mixup,manifold,smoothmix,comixup,samplepairing} mixes the
pixel intensity values of two or more images. The first two
categories are region-level augmentation, while the
third category is image-level augmentation. Within
these categories, many approaches rely entirely on
random data augmentation strategies. In contrast, others consider salient information either keeping it or deleting it in the generated image.

\begin{figure*}[!tbhp]
\centering
\includegraphics[width=1\textwidth]{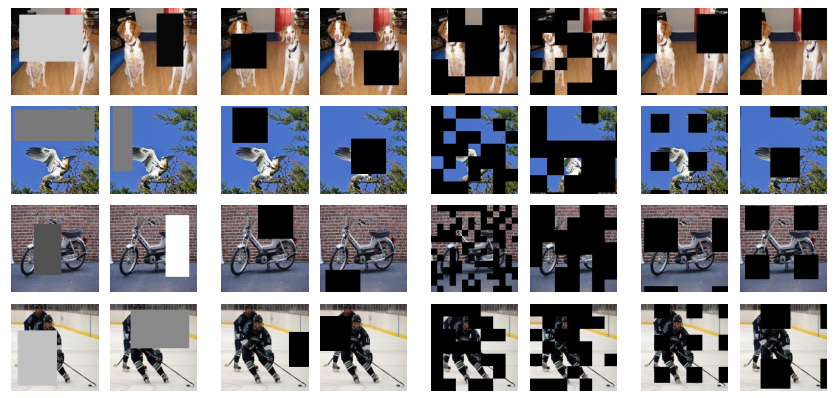}
\caption{Comparison of various cut and delete strategies, CutOut (Columns 1, 2), Random Erasing (Columns 3, 4), hide $\&$ seek (Columns 5, 6), and GridMask in the last two columns (image is taken from~\cite{gridmask}).}
\label{grid_mask_fig}
\end{figure*}

Although image mixing and deleting improves the overall performance by adding noise and ambiguity in training data and labels, it is also likely to reduce the model accuracy if performed naively. Therefore, carefully performing image mixing and deleting is essential as it should not reduce or delete the information in an image beyond a specific limit, or the model will not have sufficient information to learn and distinguish between classes. Extreme levels of changes in an image can make the label irrelevant and noisy for the model. To avoid this situation, many approaches suggested in the literature~\cite{keepaugment, attentivecutmix, puzzlemix, comixup} emphasize on retaining salient information in the augmented images.  

The generalization power of image mixing and deleting is not only limited to image classification. It has also been proven to increase accuracy for fine-grained image recognition and object detection tasks. The trained model employing suggested augmentation approaches can replace the object detection backbones, or object detection training can incorporate these augmentations; both training methods have been shown to improve accuracy over the baseline models. For fine-grained image recognition, particular augmentation strategies by image mixing are proposed. Apart from this, image mixing or deleting can efficiently be utilized along with regularization~\cite{regularization}, dropout~\cite{dropout}, or conventional image augmentations~\cite{surveyimaug} based on affine transformations, to further enhance the model performance.


This review is focused entirely on the proposed image mixing and deleting data augmentation techniques. It does not discuss any approaches that augment data using multiple geometric and color transformations~\cite{randaugment} or perform neural augmentation search~\cite{autoaugment,fastaugment}. A review of various geometric and color augmentations is available in~\cite{surveyimaug,surveyimaug2}. Moreover, we organize the image mixing and deleting-based augmentation schemes into three main categories including 1) erasing image patches, 2) cutting the image region and replacing it with some other image region, and 3) mixing pixel values of multiple images. The following sections discuss all proposed approaches within these categories in detail.

\section{Cut and Delete}
This section discusses data augmentation by deleting image patches randomly or semantically. The purpose of erasing image patches is to allow the network to learn in case of occlusions. This way, the model is forced to understand and focus on partially visible object properties. This kind of dropout is different from conventional dropout because it drops contiguous image regions, whereas values in traditional dropout work at non-contiguous locations. Here, much of the image region is provided to the network to build connections semantically among various parts of the image. Below, we would like to discuss the approaches that fall into this category.

\subsection{Cutout}
Cutout~\cite{cutout} removes constant-size square patches randomly by replacing them with any constant value. Region selection is performed by selecting a pixel location randomly and placing a square around it. In this procedure, Cutout region lying outside the borders is allowed and proven to improve performance. Patch deletion is applied on all the training images where the optimal patch size is a hyperparameter selected by performing a grid search on training data. An example is shown in Figure~\ref{grid_mask_fig}.

\subsection{Random Erasing}
Random Erasing~\cite{rerase} deletes contiguous rectangular
image regions similar to Cutout~\cite{cutout} with minor differences in the region selection procedure. Contrary to Cutout~\cite{cutout}, where deletion is applied on all the images, random erasing is performed with a probability of 0.5. 
In every iteration, region size is defined randomly with upper and lower limits on the region area and aspect ratio. The algorithm ensures erasing region stays inside the image boundaries, as opposed to Cutout. Moreover, random erasing provides region-aware deletion for object detection and person identification tasks. Regions inside the object-bounding boxes are randomly erased to generate occlusions. Figure~\ref{grid_mask_fig} shows some augmented images.

\subsection{Hide and Seek}
Hide and Seek~\cite{hidenseek} divides an image into a specified number of grids and turns each grid cell on or off with an assigned probability. This way, random small image regions that may be connected or disconnected from each other are deleted (see Figure~\ref{grid_mask_fig}). The grid cells that are turned off are replaced with the average of all the pixel values in the entire dataset.

\subsection{GridMask}
Another simple approach is GridMask~\cite{gridmask}, where the algorithm tries to overcome the drawbacks of Cutout, Random Erasing, and Hide $\&$ Seek, which are prone to deleting important information entirely or leaving it untouched without making it harder for the algorithm to learn. To handle this, the GridMask~\cite{gridmask} creates multiple blacked-out regions in evenly spaced grids to maintain a balance between deletion and retention of critical information. The number of masking grids and their sizes are tuneable. The article also compares erasing a single region vs. multiple regions, showing that GridMask is less likely to hide out foreground information completely. Even with the large box size, it retains a sufficient amount of important information for the model to learn. GridMask-generated examples are shown in Figure~\ref{grid_mask_fig}.

\subsection{Cut and Delete for Object Detection}
This section describes Cut and Delete, designed explicitly for object detection.

\vspace{2mm}
\noindent
\emph{Adversarial Spatial Dropout for Occlusion~\cite{ASDN}} drops region pixels to generate hard positives for object detection by learning key image regions. Within the proposed region, only 1/3 of pixels are dropped after sorting based on magnitudes in the heatmap. A separate network learns the salient image regions. The mask generator is penalized based on the detector's performance, similar to generative adversarial networks~\cite{GANs}.

\begin{figure}[!tbp]
\centering
\includegraphics[width=1\columnwidth]{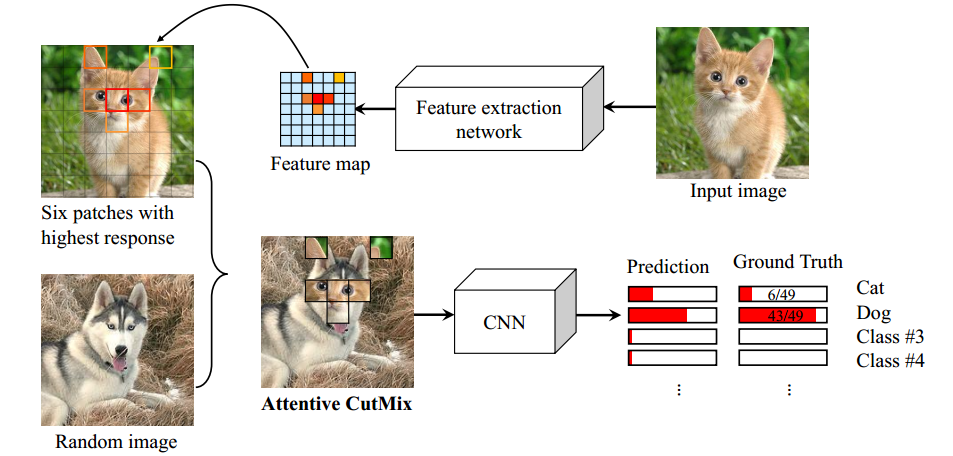}
\caption{The Attentive CutMix image augmentation procedure, attentive regions from one image are overlayed on another image, image is taken from~\cite{attentivecutmix}.}
\label{attentive_cutmix}
\end{figure}

\section{Cut and Mix}
Instead of deleting a patch, Cut and Mix replaces the patch with another image region. Thus, the image shares information coming from multiple class labels, however, the significant class label belongs to the original class label. Hence, the model learns to differentiate between two classes within a single image. This method of augmentation introduces label smoothing as a by-product.

\subsection{CutMix}
In CutMix~\cite{cutmix}, images are augmented by sampling
patch coordinates, $x, y, h, w$ from a uniform distribution. The selected patch is replaced at the corresponding location with a patch from the other randomly picked image from the current mini-batch during training. CutMix~\cite{cutmix} updates the image labels proportion to the replaced patch as defined in the Eq.~\ref{cutmix_eq}, where $M$ is the image mask, $x_a$ and $x_b$ are images, $\lambda$ is the proportion of label, and $y_a$ and $y_b$ are the labels of images.

\begin{equation}
\begin{aligned}
x_{new} = M.x_a + (1-M).x_b \\
y_{new} = \lambda.y_a + (1-\lambda).y_b
\end{aligned}
\label{cutmix_eq}
\end{equation}

\subsection{Attentive CutMix}
Attentive CutMix~\cite{attentivecutmix} builds up on CutMix. Instead of random pasting, it identifies attentive patches and pastes them at the exact location in the other image. This avoids the problem of selecting a background region that is not important for the network and updating the label information accordingly, as in Eq.~\ref{cutmix_eq}. A separate pre-trained network is employed to extract attentive regions. The attention output is mapped back onto the original image. The image is divided into a grid of patches, where six highly activated responses are pasted onto the training image as demonstrated in Figure~\ref{cutmix_images}. These image pairs are selected randomly in every training iteration. The flow diagram of Attentive CutMix is shown in Figure~\ref{attentive_cutmix}.

\begin{figure}[!tbp]
\centering
\includegraphics[width=1\columnwidth]{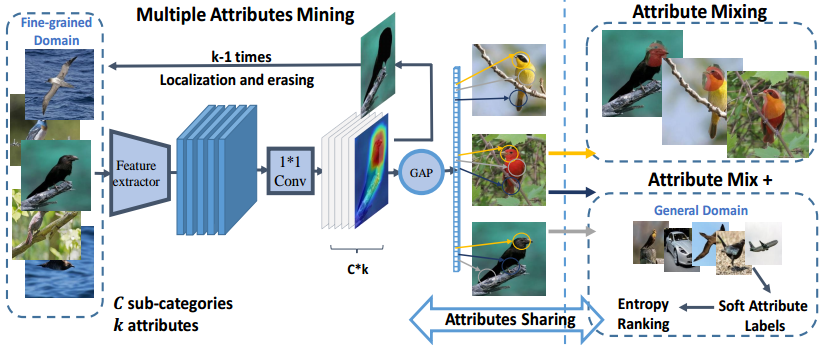}
\caption{An example of attribute mining and mixing procedure. (image is from~\cite{attributemix})}
\label{attributemix_fig}
\end{figure}

\subsection{RICAP}
Random image cropping and patching (RICAP)~\cite{ricap}
performs data augmentation by cropping four random regions from four sampled images and combining them to create a new image. The generated image has mixed labels proportional to the pasted
area. The area of cropped regions in the output
image is determined by sampling through uniform distribution. The authors RICAP~\cite{ricap} suggest three methods for common boundary points in the augmented image: anywhere-RICAP (origin can be anywhere), center-RICAP (origin can only be in the middle of the image), and corner-RICAP (origin can only be in corners). Corner-RICAP has shown the best performance due to a larger region of one image being visible to the network to learn.  


\subsection{Mixed Example}
This method~\cite{mixedexample} experimented with 14 different
types of augmentation approaches. The output image is generated using the following techniques: vertical concatenation, horizontal concatenation, mixed concatenation, random 2$\times$2, VH-mixup (vertical concatenation, horizontal concatenation, and mixup), VH-BC+ (vertical concatenation, horizontal concatenation, and between-class), random square, random column interval, random row interval, random rows, random columns, random pixels, random elements, and noisy mixup. From all these approaches, VHmixup has the best performance.

\subsection{CowMask}
CowMask~\cite{cowmask} is used in semi-supervised learning
where original and augmented images are brought
closer during training. CowMask suggests two types
of mixing approaches 1) erasing and 2) mixing two
images similar to CutMix. This technique's mask is irregular rather than rectangular, generated by applying a Gaussian filter (scale $\sigma$) on a normally distributed noise image. A suitable threshold is selected to ensure a proportion $p$ of non-masked image pixels are present in the output image. Pixel values below the threshold are either erased or replaced by the pixel values of the noise image at the corresponding locations.

A very similar masking approach for data augmentation is FMix~\cite{fmix} which uses the inverse Fourier transform of a noise image to generate binary masks containing top $\lambda wh$ pixels with 1 values.     

\subsection{ResizeMix}
ResizeMix~\cite{resizemix} performs random image cropping
and pasting. It empirically proves to perform better than salient and other non-salient image mixing methods, as shown in Table~\ref{img_clf_tab}. ResizeMix solves the random region cropping problem that misallocates the output image label in some instances where the pasted region does not contain any object information. The labels of the output image are updated as per Eq.~\ref{cutmix_eq}. ResizeMix handles this by completely scaling down (scale rate is sampled from a uniform distribution) the selected image and pasting it randomly on the target image, as shown in Figure~\ref{cutmix_images}.

\subsection{SaliencyMix} 
Instead of random Cut and Mix, SaliencyMix~\cite{saliencymix} extracts the salient region and pastes them on the corresponding location in the target image. The salient region is extracted around the maximum intensity pixel location in the saliency map. Augmented image labels are updated as per Eq.~\ref{cutmix_eq}.

\subsection{KeepAugment}
Similar to SaliencyMix~\cite{saliencymix} and Attentive CutMix~\cite{attentivecutmix}, KeepAugment~\cite{keepaugment} performs augmentation based on the 
salient region. However, this approach uses one image for the augmentation. KeepAugment identifies the salient area in an image and ensures that the image generated by the augmentation strategies, for example, Cutout~\cite{cutout}, randaugment~\cite{randaugment}, CutMix~\cite{cutmix} or autoaugment~\cite{autoaugment}, contains the salient region in it.

\subsection{RecursiveMix}
RecursiveMix~\cite{recursivemix} iteratively mixes images from the previous iteration into the current iteration. Similar to ResizeMix~\cite{resizemix}, it resizes and pastes the mixing image on top of the input image but employs previously generated output (containing multiple classes), mixing recursively while maintaining the history of one mini-batch. Labels of the generated image have class values proportional to the amount of the region. The network is trained with classification and consistency loss to learn semantic representations in multi-scale and spatial-variant views, as given in Eq.~\ref{recursive_mix_eq}
\begin{equation}
\begin{aligned}
L = L_{CE}(x_m, y_m) + \omega \lambda L_{KL}(p_{roi}, p^h),
\end{aligned}
\label{recursive_mix_eq}
\end{equation}
where $CE$ is cross-entropy loss, $x_m$ and $y_m$ are the mixed image and label, respectively, $\omega$ is the consistency loss weight, $\lambda$ is the image mixing ratio, $KL$ is the Kullback Leibler Divergence, $p_{roi}$ is the network's prediction for the resized smaller pasted image on the input image, whose features are extracted and aligned by $roi$, \ie, RoIAlign operator~\cite{mask_rcnn}, and $p^h$ is the network's prediction for the historical image. The KL divergence minimizes the model's prediction variance between the historical and generated images. Examples of the RecursiveMix generated images are shown in Figure~\ref{cutmix_images}.

\subsection{LUMix}
LUMix~\cite{lumix} corrects the label misallocation of CutMix, perturbing the labels by adding noise and reducing the input class label value if the output image does not contain the class information. The label equation in Eq.~\ref{cutmix_eq} updates to the following:
\begin{equation}
\begin{aligned}
\lambda = (1 - r_1 - r_2)\lambda_0 + r_1\lambda_r + r_2\lambda_s,   
\end{aligned}
\end{equation}
where $r_1$ and $r_2$ are the hyper-parameters, $\lambda_0$ is the CutMix $\lambda$ as in Eq.~\ref{cutmix_eq}, $\lambda_r$ is a random value drawn from the Beta distribution, and $\lambda_s$ is calculated in Eq.~\ref{lumix_lambda_s} using the network's predictions on the CutMix input image.
\begin{equation}
\begin{aligned}
\lambda_s = \frac{p_{y_a}}{p_{y_a} + p_{y_b}}.
\end{aligned}
\label{lumix_lambda_s}
\end{equation}

In the above Eq., $p_{y_a}$ and $p_{y_b}$ are the predicted probabilities for the $y_a$ and $y_b$ classes, respectively. The $\lambda$ in Eq.~\ref{lumix_lambda_s} updates labels in Eq.~\ref{cutmix_eq}, later used for backpropagation. The lower value of $p_{y_a}$ decreases the $\lambda$ value, which in turn reduces the penalization for the $y_a$ class, suggesting the absence of input class information.

\subsection{Saliency Grafting}
\begin{figure*}[!tbhp]

\begin{center}
\begin{tabular}{c@{ } c@{ } c@{ } c@{ } c@{ } c@{ } c}
   & \multicolumn{4}{c}{Random Images} &  Input Image \\

  &\includegraphics[width=.12\textwidth]{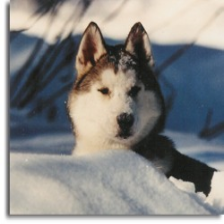}& 
  \includegraphics[width=.12\textwidth]{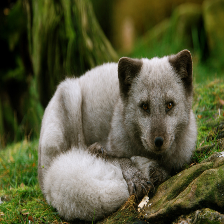}&
  \includegraphics[width=.12\textwidth]{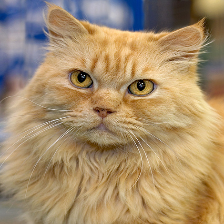}&
  \includegraphics[width=.12\textwidth]{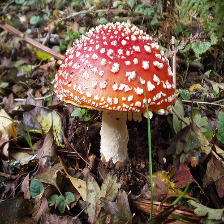}&
  \includegraphics[width=.12\textwidth]{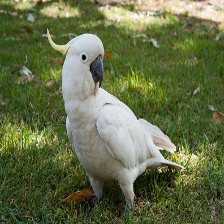}\\

   \raisebox{.5\height}{\rotatebox{90}{CutMix}} &
  \includegraphics[width=.12\textwidth]{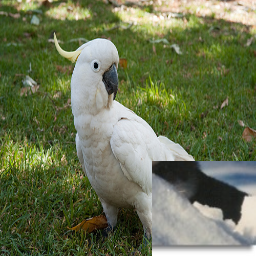}&
  \includegraphics[width=.12\textwidth]{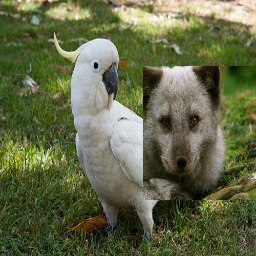}&  
  \includegraphics[width=.12\textwidth]{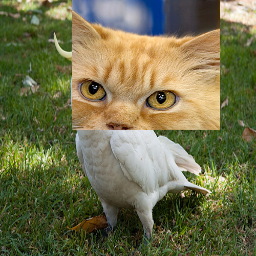}&
  \includegraphics[width=.12\textwidth]{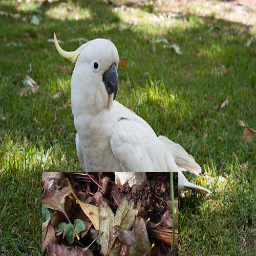}\\

  \rotatebox{90}{Att. CutMix}&
  \includegraphics[width=.12\textwidth]{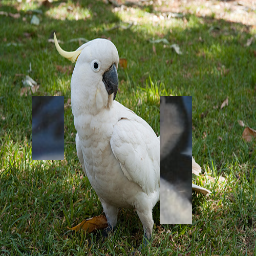}&
  \includegraphics[width=.12\textwidth]{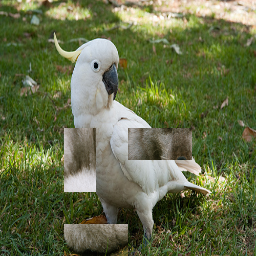}&  
  \includegraphics[width=.12\textwidth]{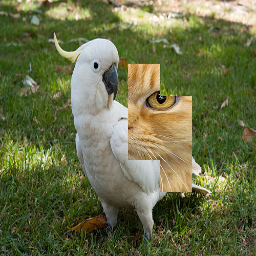}&
  \includegraphics[width=.12\textwidth]{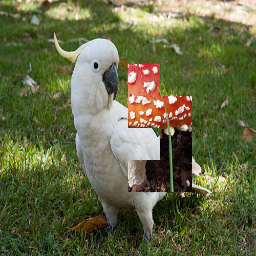}\\

  \raisebox{.2\height}{\rotatebox{90}{CowMask}}&
  \includegraphics[width=.12\textwidth]{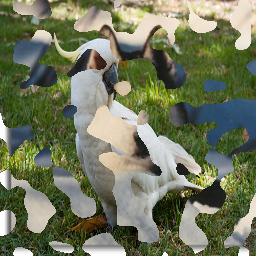}&
  \includegraphics[width=.12\textwidth]{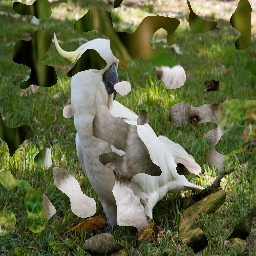}&  
  \includegraphics[width=.12\textwidth]{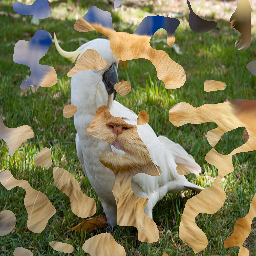}&
  \includegraphics[width=.12\textwidth]{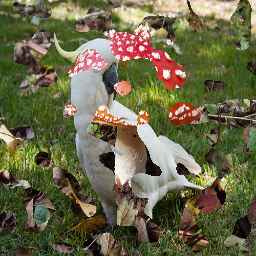}\\

  \raisebox{.2\height}{\rotatebox{90}{ResizeMix}}&
  \includegraphics[width=.12\textwidth]{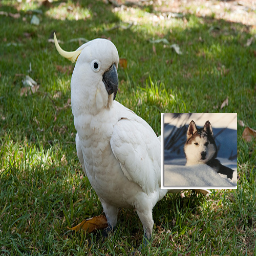}&
  \includegraphics[width=.12\textwidth]{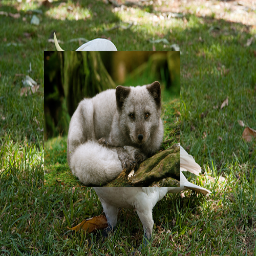}&  
  \includegraphics[width=.12\textwidth]{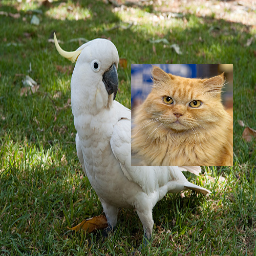}&
  \includegraphics[width=.12\textwidth]{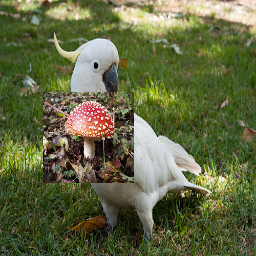}\\

  \raisebox{.15\height}{\rotatebox{90}{KeepAug.}}&
  \includegraphics[width=.12\textwidth]{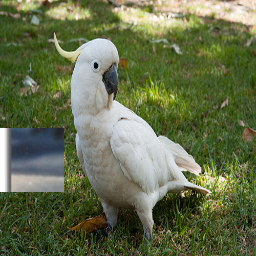}&
  \includegraphics[width=.12\textwidth]{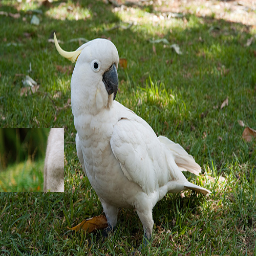}&  
  \includegraphics[width=.12\textwidth]{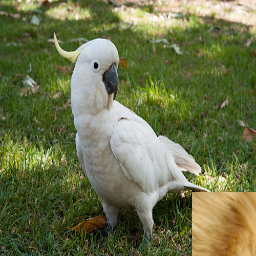}&
  \includegraphics[width=.12\textwidth]{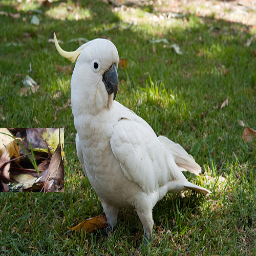}\\

  \raisebox{.1\height}{\rotatebox{90}{SaliencyMix}}&
  \includegraphics[width=.12\textwidth]{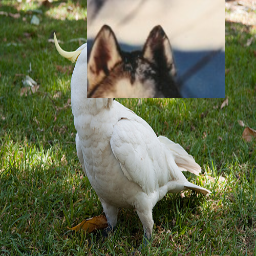}&
  \includegraphics[width=.12\textwidth]{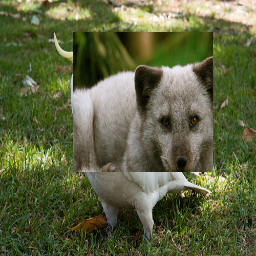}&  
  \includegraphics[width=.12\textwidth]{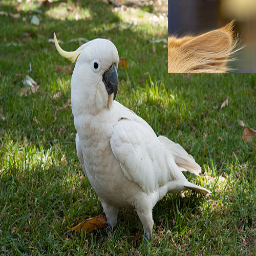}&
  \includegraphics[width=.12\textwidth]{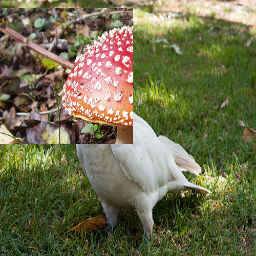}\\

  \raisebox{.01\height}{\rotatebox{90}{RecursiveMix}}&
  \includegraphics[width=.12\textwidth]{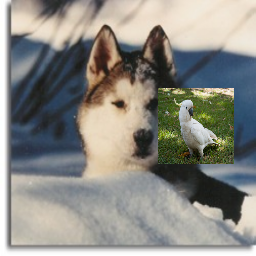}&
  \includegraphics[width=.12\textwidth]{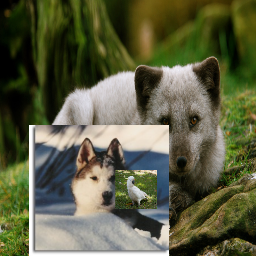}&  
  \includegraphics[width=.12\textwidth]{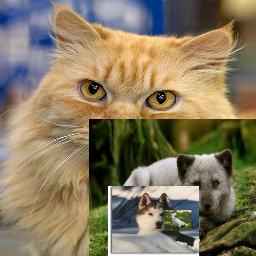}&
  \includegraphics[width=.12\textwidth]{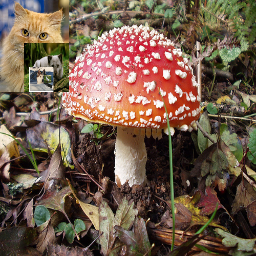}\\

  \raisebox{.05\height}{\rotatebox{90}{SaliencyG.}}&
  \includegraphics[width=.12\textwidth]{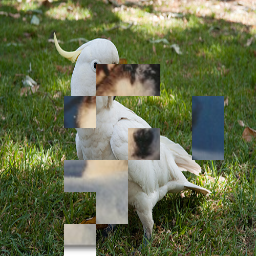}&
  \includegraphics[width=.12\textwidth]{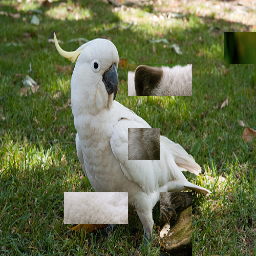}&  
  \includegraphics[width=.12\textwidth]{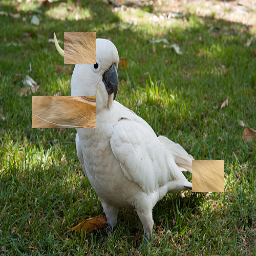}&
  \includegraphics[width=.12\textwidth]{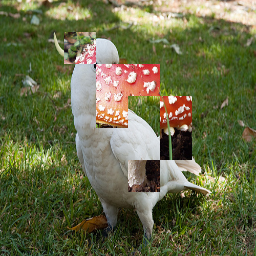}\\
\end{tabular}
\end{center}
\caption{A comparison of various cut and mix approaches. The topmost row shows original images, while the other rows are augmented images. SaliencyG. refers to Saliency Grafting~\cite{saliency_grafting}.}
\label{cutmix_images}
\end{figure*}

Saliency Grafting~\cite{saliency_grafting} incorporates salient information for both the image and label mixing. Instead of pasting the most salient patches from the random image, Saliency Grafting selects patches randomly using a 2D-Bernoulli distribution. The labels are modified based on the amount of salient information of both images present in the output image. The following Eq.~\ref{saliency_grafting_eq} describes the label-mixing procedure: 
\begin{equation}
\begin{aligned}
y = \lambda(S_a, S_b, M)y_a + (1 - \lambda(S_a, S_b, M))y_b, \\
\end{aligned}
\label{saliency_grafting_eq}
\end{equation}
where 
$\lambda = \frac{I(S_a, M)}{I(S_a, M) + I(S_b, 1-M)}$
 and $I = \frac{||S\circ M||_2}{||S||_2}$. Moreover, $S$ is the attention map, $M$ is the mask, $I$ is the salient information present in the output image, $\lambda$ is the hyper-parameter and $y$, $y_a$, and $y_b$ are the image labels. See examples of the Saliency Grafting augmented images in Figure~\ref{cutmix_images}.  

\subsection{Fine-Grained Image Recognition}
\vspace{2mm}
\noindent
\emph{Attribute Mix~\cite{attributemix}} augments images based on semantically extracted image attributes. It trains an attribute classifier by extracting $k$ attributes (\eg, leg, head, and wings of a bird) from each image. The attribute mining procedure for every image is performed  $k$ times repetitively. For each iteration, an attribute is masked out from the original image based on the most discriminative region in the attention map. With these images, an attribute-level classifier is trained to generate new images for the actual classification model. For the given two images, the attribute-level classifier identifies attribute masks; these masks are randomly picked to create a new training image. The procedure of label update is the same as in Eq.~\ref{cutmix_eq}. The attribute mining and attribute mixing process can be seen in Figure~\ref{attributemix_fig}.

\begin{figure}[!tbp]
\centering
\includegraphics[width=1\columnwidth]{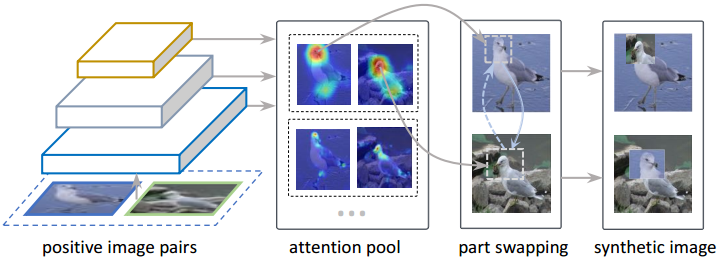}
\caption{Architecture of the Intra-class part swapping. The figure is from~\cite{intraPartSwap}.}
\label{partswap_fig}
\end{figure}

\vspace{2mm}
\noindent
\emph{Intra-Class Part Swapping~\cite{intraPartSwap}} replaces the most attentive regions, extracted using a classification activation map (CAM) thresholded for the most prominent region, of one image with another. The attentive region in the source image is scaled and translated according to the attentive region of the target image for region replacement. The label information of the output is similar to the target image as this approach relies on augmenting similar class images. Figure~\ref{partswap_fig} presents the architecture of intra-class part swapping.

\begin{figure}[!tbp]
\centering
\includegraphics[height=5cm, width=1\columnwidth]{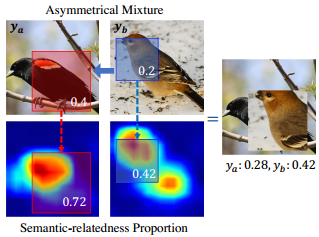}
\caption{Example of SnapMix taken from~\cite{snapmix}}
\label{snapmix_fig}
\end{figure}

\vspace{2mm}
\noindent
\emph{SnapMix~\cite{snapmix}} augments training images by extracting and merging random image regions of different sizes, where the region size for both images is drawn through the beta distribution. Generated image label is assigned based on semantic composition from normalized (sum to one) CAMs. Figure~\ref{snapmix_fig} illustrates an example of image augmentation. The procedure of label generation is similar to Eq.~\ref{cutmix_eq}. However, the summation of label coefficients can exceed one depending on the semantic composition of the output image.  

\subsection{Cut and Mix for Transformers}
\vspace{2mm}
\noindent
\emph{TransMix~\cite{transmix}} employs Vision Transformers (ViTs)~\cite{vits} to identify attentive regions taken from the random image in CutMix generated image. These attentive regions are used to update labels ($\lambda$ in Eq.~\ref{cutmix_eq}) based on the information present in the output image.

\vspace{2mm}
\noindent\emph{TokenMix~\cite{tokenmix}} splits the mask into multiple patches, where patches are on and off at distributed locations instead of one large rectangular region, as in CutMix. The labels are updated according to the attentive regions CutMix in the output image.

\subsection{Cut and Mix for Object Detection}
This section discusses Cut and Mix approaches specifically designed for object detection.

\vspace{2mm}
\noindent
\emph{Visual Context Augmentation~\cite{VCAugment1}.}
This augmentation strategy learns to place object instances at an image location depending on the surrounding context. A neural network is trained for this purpose. The training data is prepared to generate a context image with the masked-out object inside it. From an image, 200 context sub-images are generated surrounding the blacked-out bounding box. The neural network learns to predict the category (object or background) in masked pixels. During testing, the context network identifies plausible object bounding boxes. The object instances are placed inside the selected boxes to generate a new training image. These newly created images are used as training data for the other network, for example, object detectors.

\vspace{2mm}
\noindent
\emph{Cut, Paste and Learn ~\cite{cutpasteandlearn}.}
This simple approach generates new data by extracting object instances and pasting them on randomly selected background images. Instances are blended with various blending approaches, for example, Gaussian blurring and Poison blending, to reduce pixel artifacts around the augmented object boundaries. Added instances are also rotated, occluded, and truncated to make the learning algorithm robust. This simple data mixing approach improves object detection performance.

\section{Mix and Up}
This type of data mixing is different from Cutout
and CutMix, in the sense that it mixes pixel values of two
images instead of cropping or erasing image regions.
Mix and Up regularizes neural networks by forcing them
to learn linear interpolations between training images. The network trained with Mix and Up has shown robustness to adversarial attacks with improved performance on the test set. 

\subsection{MixUp}
MixUp~\cite{mixup} generates a new image by mixing pixel
values of two randomly selected images. The mixing factor decides the proportion of pixel strength during data mixing $\lambda$ $\in$ [0-1] as shown in Eq.~\ref{mixup_eq}. The values for $\lambda$ are drawn from the beta distribution.

\begin{equation}
\begin{aligned}
x_{new} = \lambda.x_a + (1-\lambda).x_b \\
y_{new} = \lambda.y_a + (1-\lambda).y_b.
\end{aligned}
\label{mixup_eq}
\end{equation}

\vspace{2mm}
\noindent
\emph{RegMixUp~\cite{regmixup}} employs MixUp as a regularizer along with the standard cross-entropy loss.

\subsection{Manifold MixUp}
Different from MixUp, Manifold MixUp~\cite{manifold} mixes feature values generated from intermediate neural network layers. For two random images, inputs are fed forward up to the $k$ layer of the network, where the output feature maps are mixed based on Eq.~\ref{mixup_eq}. The mixed feature maps are inputted to the next layer and forward propagated up to the last layer. Finally, backward propagation is performed in the standard way with updated labels (similar to MixUp).

\vspace{2mm}
\noindent
\emph{Noisy Feature MixUp~\cite{noisyfeaturemixup}.} This method injects additive and multiplicative noise into the Manifold MixUp output. Eq.~\ref{mixup_eq} for Noisy Feature MixUp becomes:
\begin{equation}
\begin{aligned}
x_{new} = (1 + \sigma_1\xi_1)(\lambda.x_a + (1-\lambda).x_b) + \sigma_2\xi_2, \\
\end{aligned}
\label{noisy_feature_mixup_eq}    
\end{equation}
where $\sigma$ is a scalar, and $\xi$ is drawn from a probability distribution (Gaussian) and has the same dimension as the mixing output image. 

\begin{figure}[!tbp]
\centering
\includegraphics[width=1\columnwidth]{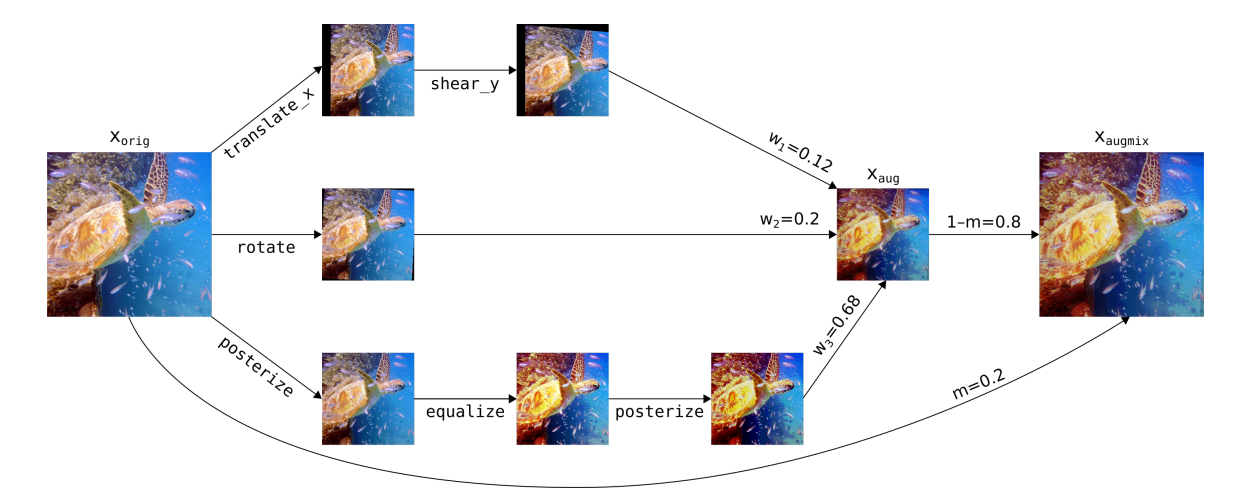}
\caption{AugMix~\cite{augmix} data augmentation procedure. Multiple transformations are applied randomly on the image in separate branches that are later combined to generate one image. Example taken from~\cite{augmix}.}
\label{augmix_fig}
\end{figure}

\subsection{AugMix}
AugMix~\cite{augmix} performs data mixing using the input
image itself. It transforms (translate, shear, rotate, \etc) the input image and mixes it with the original image. Image transformation involves a series of randomly selected augmentation operations applied with three parallel augmentation chains. Each chain has a composition of functions that could include employing, for example, translation on input image followed by shear, \etc. The output of these three chains is three images mixed to form a new image (see Figure~\ref{augmix_fig}). This new image is later mixed with the original image to generate the final augmented output image, as per Eq.~\ref{mixup_eq}. The transformations in AugMix~\cite{augmix}  are selected from autoaugment~\cite{autoaugment}. A comparison of AugMix~\cite{augmix} augmented images with other methodologies is given in Figure~\ref{mixup_images}.

\begin{figure}[!tbp]
\centering
\includegraphics[width=1\columnwidth]{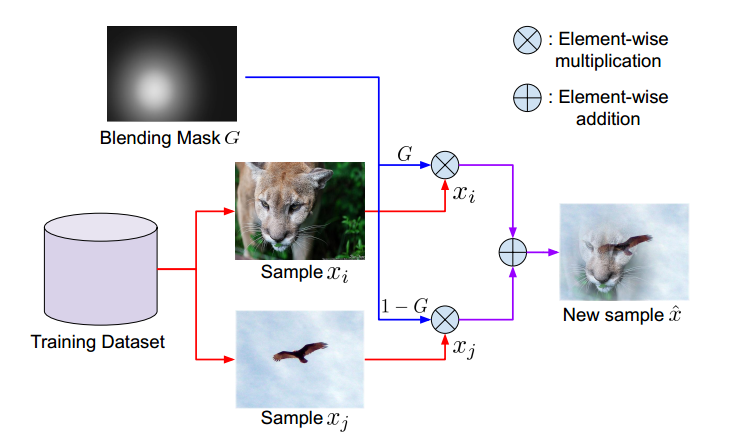}
\caption{An example of network training via SmoothMix
strategy, where the figure is from~\cite{smoothmix}.}
\label{smoothmix_fig}
\end{figure}

\subsection{SmoothMix}
SmoothMix~\cite{smoothmix} is a mask-based approach matching closely with the Cutout and CutMix techniques. However, it has a few differences: 1) the mask has soft edges with gradually decreasing intensity, and 2) the mixing strategy is the same as in Eq.~\ref{mixup_eq}. The augmented image has mixed pixel values depending on the strength of the mask, as shown in  Figure~\ref{smoothmix_fig}.
The authors suggest various square and circular
masks containing smooth edges. From these, circular masks generated using Gaussian distribution perform better. The following decides the value of $\lambda$. 

\begin{equation}
\lambda = \frac{\sum_{i=1}^{W}\sum_{j=1}^{H}G_{ij}}{WH},
\label{smooth_mix_eq}
\end{equation}
where $G_{ij}$ is the pixel value of mask $G$, $H$ is height,
and $W$ is width. With the given mask, the equation
to mix two images becomes 
\begin{equation}
\begin{aligned}
x_{new} = G.x_a + (1-G).x_b \\
y_{new} = \lambda.y_a + (1-\lambda).y_b.
\end{aligned}
\end{equation}

\subsection{Co-Mixup}
Co-Mixup~\cite{comixup} performs salient image mixing on a batch of input images to generate a batch of augmented images. This technique maximizes saliency in output images by penalizations to ensure local data
smoothness and diverse image regions. Examples of generated images by Co-Mixup are shown in  Figure~\ref{mixup_images}.

\subsection{Sample Pairing}
This technique~\cite{samplepairing} merges two images by averaging their pixel intensities. The resultant image has
the same training image label as opposed to MixUp and other approaches where labels are updated according
to the proportion of image mixing.

\subsection{Puzzle Mix}
Puzzle Mix~\cite{puzzlemix}, as shown in Figure~\ref{mixup_images}, learns to augment two images optimally based on saliency. The algorithm employs the following procedure 1) images are divided into regions for the MixUp, and 2) the algorithm learns to transport the salient region of one image such that the output image has the maximized saliency from both images.

\subsection{SuperMix}
SuperMix~\cite{supermix} performs MixUp~\cite{mixup} on multiple images. It mixes salient information extracted using a set of mixing masks $M = \{m_i\}_{i=0}^k$. These masks are optimized to generate an output image that contains the label information equivalent to the predictions of a teacher model, given below: 
\begin{equation}
\begin{aligned}
\hat{x} = \sum_{i=0}^{k-1}m_ix_i  \quad \text{and} \quad 
\hat{y} = \sum_{i=0}^{k-1}r_i\delta(y^T(x_i)), 
\end{aligned}
\label{supermix_eq}    
\end{equation}
where $k$, $r$, $\delta$, and $y^T(x_i)$ are the mixing size, random sample from Dirichlet distribution,  one hot encoding, and the teacher network prediction for the $ith$ image, respectively. The masks are optimized by minimizing the KL divergence between the teacher network predictions for the mixed image and the labels (from Eq.~\ref{supermix_eq}) defined as 
\begin{equation}
\begin{aligned}
\underset{i=0,...,k-1}{min} \ KL(f^T(\hat{x})\parallel\hat{y}) + \lambda_{\sigma}L_{\sigma}(m_i) 
+ \lambda_sL_s(m_i). \\
\end{aligned}
\label{supermix_opt_eq}    
\end{equation}

\begin{figure}[!tbp]
\centering
\includegraphics[width=1\columnwidth]{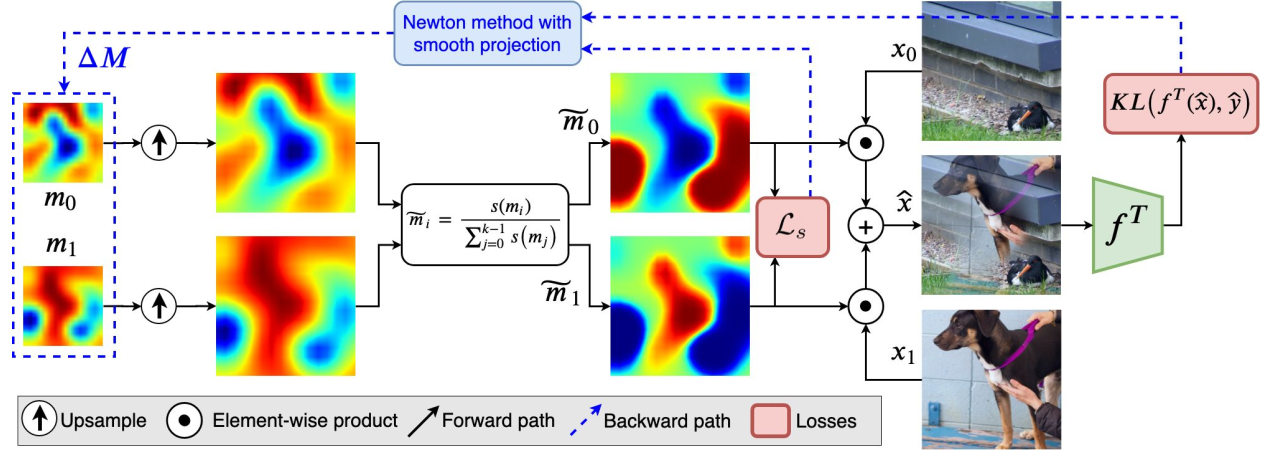}
\caption{SuperMix augmented image generation process, the figure is taken from~\cite{supermix}.}
\label{supermix_fig}
\end{figure}

\begin{figure*}[tbhp]
\begin{center}
\begin{tabular}{c@{ } c@{ } c@{ } c@{ } c@{ } c@{ } c}
   &\multicolumn{4}{c}{Input Images} &  Random Image \\
  
  &\includegraphics[width=.14\textwidth]{mix_inp2}&  
  \includegraphics[width=.14\textwidth]{mix_inp3}&
  \includegraphics[width=.14\textwidth]{mix_inp4}&
  \includegraphics[width=.14\textwidth]{mix_inp5}&
  \includegraphics[width=.14\textwidth]{mix_inp1}\\

  \raisebox{.5\height}{\rotatebox{90}{MixUp}}&
  \includegraphics[width=.14\textwidth]{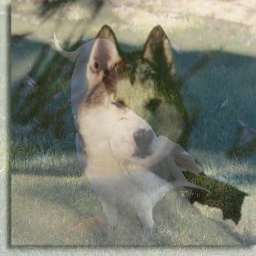}&
  \includegraphics[width=.14\textwidth]{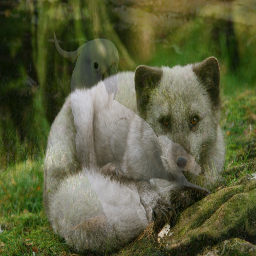}&  
  \includegraphics[width=.14\textwidth]{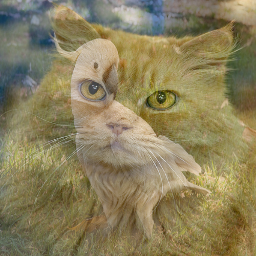}&
  \includegraphics[width=.14\textwidth]{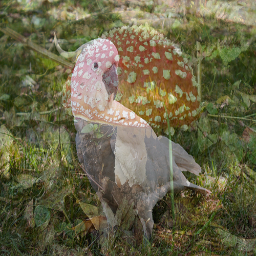}\\

  \raisebox{.5\height}{\rotatebox{90}{AugMix}} &
  \includegraphics[width=.14\textwidth]{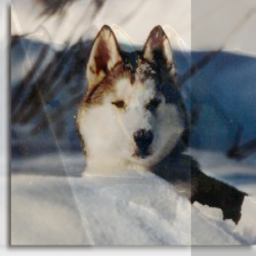}&
  \includegraphics[width=.14\textwidth]{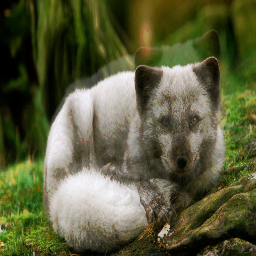}&  
  \includegraphics[width=.14\textwidth]{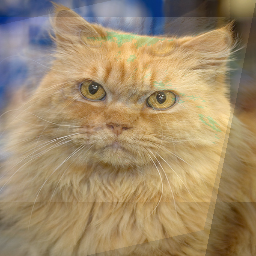}&
  \includegraphics[width=.14\textwidth]{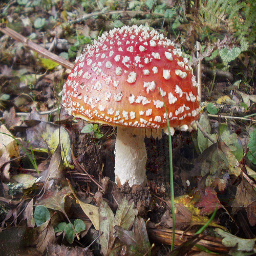}\\

  \raisebox{.2\height}{\rotatebox{90}{SmoothMix}} &
  \includegraphics[width=.14\textwidth]{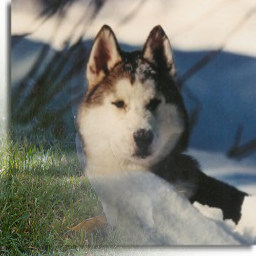}&
  \includegraphics[width=.14\textwidth]{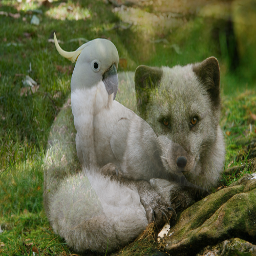}&  
  \includegraphics[width=.14\textwidth]{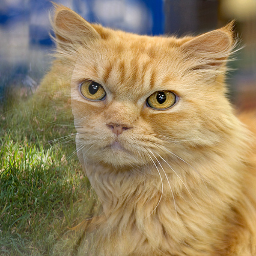}&
  \includegraphics[width=.14\textwidth]{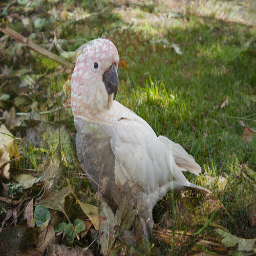}\\

  \raisebox{.2\height}{\rotatebox{90}{Co-MixUp}}& 
  \includegraphics[width=.14\textwidth]{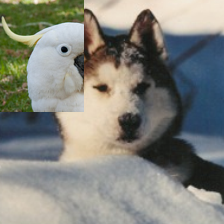}&
  \includegraphics[width=.14\textwidth]{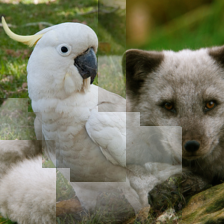}&  
  \includegraphics[width=.14\textwidth]{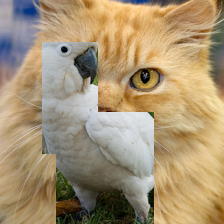}&
  \includegraphics[width=.14\textwidth]{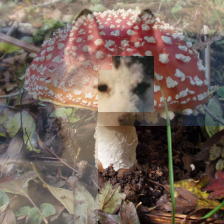}\\

 \raisebox{.2\height}{\rotatebox{90}{Puzzle Mix}}& 
  \includegraphics[width=.14\textwidth]{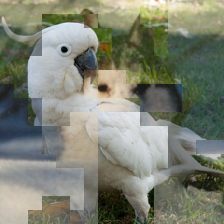}&
  \includegraphics[width=.14\textwidth]{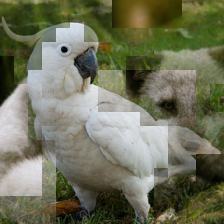}&  
  \includegraphics[width=.14\textwidth]{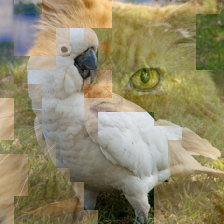}&
  \includegraphics[width=.14\textwidth]{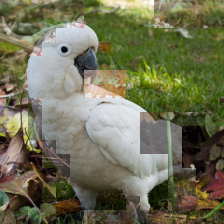}\\
  
  \raisebox{.35\height}{\rotatebox{90}{SuperMix}}&
  \includegraphics[width=.14\textwidth]{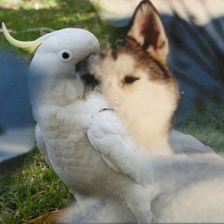}&
  \includegraphics[width=.14\textwidth]{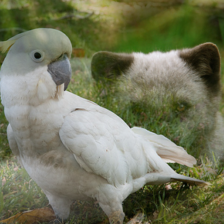}&  
  \includegraphics[width=.14\textwidth]{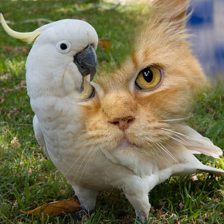}&
  \includegraphics[width=.14\textwidth]{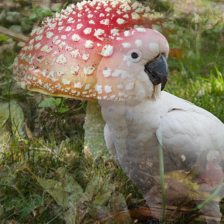}\\
    
  \raisebox{.5\height}{\rotatebox{90}{PixMix}}&
  \includegraphics[width=.14\textwidth]{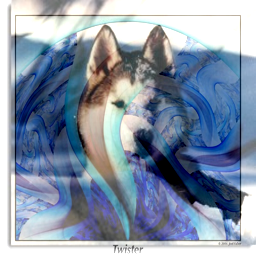}&
  \includegraphics[width=.14\textwidth]{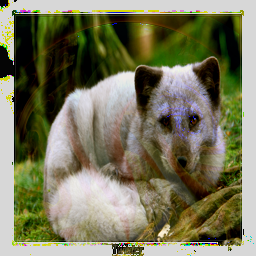}&  
  \includegraphics[width=.14\textwidth]{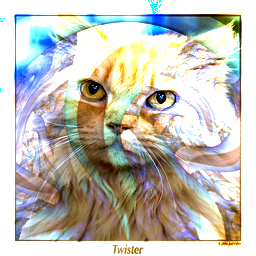}&
  \includegraphics[width=.14\textwidth]{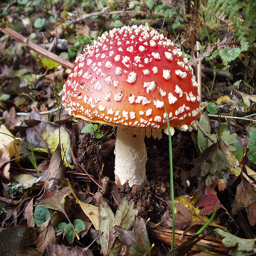}\\
\end{tabular}
\end{center}
\caption{A comparison of various MixUp techniques. The topmost row shows original images, whereas other rows are augmented images.}
\label{mixup_images}
\end{figure*}

Here, $f^T$ is the teacher network, $\lambda_{\sigma}$ $\&$ $\lambda_s$ are hyper-parameters, $L_{\sigma}$ is the total variation norm to penalize mask roughness, and $L_s$ encourages the network to have mask values approaching to 0 or 1, where $L_s = \frac{1}{kWH}\sum_{i}m_i(m_i - 1)$. Figure~\ref{supermix_fig} illustrates this optimization process and the classification model that uses the resultant output image from the optimization process for training; examples are shown in Figure~\ref{mixup_images}.

\subsection{AutoMix}
\begin{figure}[!tbp]
\centering
\includegraphics[width=1\columnwidth]{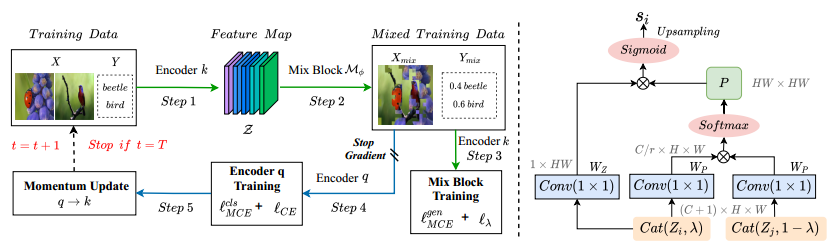}
\caption{AutoMix image mixing and training procedure, the figure is taken from~\cite{automix}.}
\label{automix_fig}
\end{figure}

AutoMix~\cite{automix} employs a mix block network to generate image mixing masks to avoid label mismatch in MixUp. Mix block is a cross-attention module that takes mixing images feature maps as input, extracted from a momentum encoder. Mix block, momentum encoder, and the classification model are trained end-to-end with a momentum pipeline, where the momentum encoder is the exponential moving average of the classification model. The complete training pipeline is visible in Figure~\ref{automix_fig}. A joint loss, given in Eq.~\ref{automix_eq}, is optimized during the training process.
\begin{equation}
\begin{aligned}
l_{\lambda} = \gamma max(&||\lambda - \frac{1}{HW} \sum_{h, w}{s_a}|| - \epsilon, 0) \\
L &= l_{ce} + l_{\lambda} .
\end{aligned}
\label{automix_eq}    
\end{equation}

In the above Eq., $l_{\lambda}$ is the mix block loss, $\gamma$, $\lambda$, $\epsilon$ are hyper-parameters, $s_a$ is the mask for the $x_a$ image, $l_{ce}$ is the classification and mask generation cross-entropy loss, and $L$ is the combined loss. 

\subsection{ReMix}
ReMix~\cite{remix} addresses the issue of class imbalance by
generating mixed images for minority classes. The process of remixing is identical to Eq.~\ref{mixup_eq} for data mixing. However, in the case of label assignment, it sets the label of the output image to the minority class.

\subsection{PixMix}
\begin{figure}[!tbp]
\centering
\includegraphics[width=1\columnwidth]{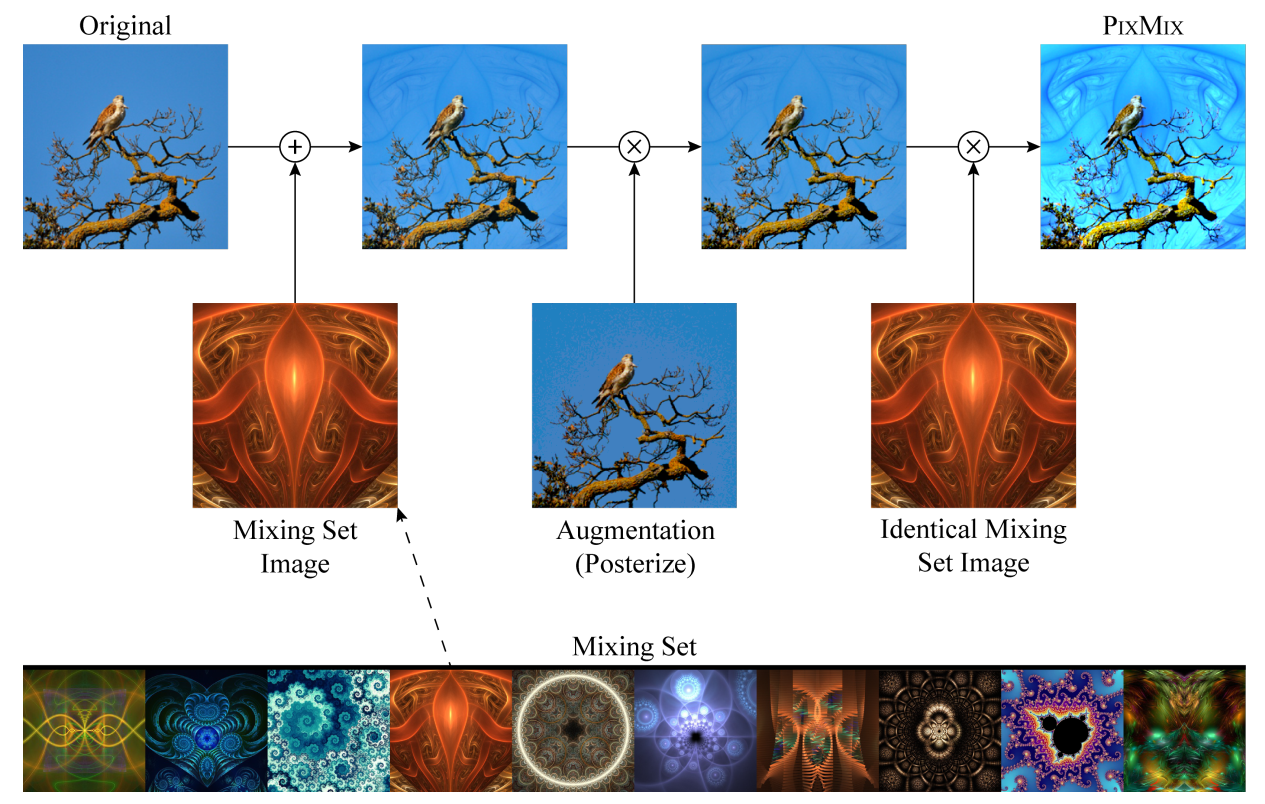}
\caption{Image augmentation strategy of PixMix. The bottom row shows the mixing set fractals, where the image is taken from~\cite{pixmix}.}
\label{pixmix_fig}
\end{figure}

PixMix~\cite{pixmix} improves the model performance in various dimensions, robustness, consistency, calibration, corruption, adversarial, and anomaly detection while maintaining an excellent comparable accuracy to other methods on a clean dataset. The mixing procedure employs fractals and feature map visualizations to mix with the input image, rather than some random dataset image as in other Mix $\&$ Up methods, depicted in  Figure~\ref{pixmix_fig}. The input image is augmented $k$ times, either by addition or multiplication. Within these augmentation rounds, input image mixes with fractals, feature map visualizations, and augmented (rotated, posterized, solarized, \etc) versions of the original image.

\subsection{StyleMix}
\begin{figure}[!tbp]
\centering
\includegraphics[width=1\columnwidth]{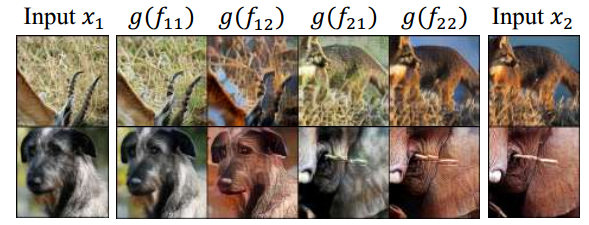}
\caption{A pre-trained encoder and decoder generate StyleMix images. We can observe that $g(f_{11})$ has the content and style of $x_1$. Similarly, $g(f_{12})$ contains content of $x_1$ and style of $x_2$ and so on. Image is from~\cite{stylemix}.}
\label{stylemix_fig}
\end{figure}

StyleMix~\citep{stylemix} generates new images by mixing style and content from image mixing pairs; an example is available in  Figure~\ref{stylemix_fig}. The style of one image is transferred to another using adaptive instance normalization~\cite{adain} on encoded feature vectors. Lets consider two images $x_1$ and $x_2$, their encoded and adaptive instance normalized~\cite{adain} feature vectors are following:

\begin{equation}
\begin{aligned}
f_{11} &= f(x_1) \\
f_{22} &= f(x_2) \\
f_{12} &= AdaIN(f_{11}, f_{22}) \\
f_{21} &= AdaIN(f_{22}, f_{11}),
\end{aligned}
\label{stylemix_feat_vec_eq}
\end{equation}
where $f$ is a VGG network up to the first few layers. The $AdaIN(f_{12}, f_{21})$ is from~\cite{adain} and given by:
\begin{equation}
\begin{aligned}
AdaIN(f_{ii}, f_{jj}) = \sigma(f_{ii})\frac{f_{ii} - \mu(f_{ii})}{\sigma(f_{ii})} + \mu(f_{jj}).\\
\end{aligned}
\label{adain_eq}
\end{equation}

The mean $\mu$ and variance $\sigma$ are computed individually for each channel. The mixed image and its label are generated by Eq.~\ref{stylemix_eq} below.

\begin{equation}
\begin{aligned}
x_m &= g(tf_{11} + (1-r_c - r_s +t)f_{22} \\
&+ (r_c - t)f_{12} + (r_s - t)f_{21})), \\
y_m &= ry_c + (1 - r)y_s, \\
y_c &= r_cy_1 + (1 - r_c)y_2, \\
y_s &= r_sy_1 + (1 - r_s)y_2,
\end{aligned}
\label{stylemix_eq}
\end{equation}
where $r_c$ and $r_s$ are content and style parameters, respectively. The interpolated image $x_m$ contains $r_c$ content of $x_1$ ($1 - r_c$ of $x_2$) and $r_s$ style of $x_2$ ($1 - r_s$ of $x_1$), whereas $t$ is a free parameter within the range $max(0, r_c + r_s -1) \leq t \leq min(r_c, r_s)$.  

In addition to MixUp-based style mixing, StyleMix also suggests StyleCutMix, where the output image has the styles either from the cut-pasted region of the random image or the original image depending on the parameters in Eq.~\ref{stylemix_eq}. 

\subsection{AlignMixUp}
\begin{figure*}[t]
\centering
\includegraphics[width=0.6\textwidth]{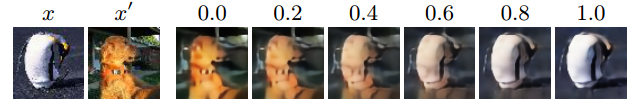}
\caption{An example of AlignMixUp from~\cite{alignmixup} shows interpolated and aligned images generated through the decoder. The mixing ratio values $[0, 1]$ are shown with their effects on the output image.}
\label{alignmixup_fig}
\end{figure*}

AlignMixUp~\cite{alignmixup} geometrically aligns and interpolates two images in the feature space. The reconstruction of aligned-mixup feature space displays the content of one image and the texture of the other; an example is shown in  Figure~\ref{alignmixup_fig}. AlignMixUp uses sinkhorn~\cite{sinkhorn} algorithm to align features, and MixUp with interpolating original and aligned feature spaces of two images. 

\subsection{Mix and Up for Transformers}
\vspace{2mm}
\noindent
\emph{TokenMixup~\cite{tokenmixup}.} In this method, images in a mini-batch are paired optimally to increase salient information in the output. TokenMixup calculates the saliency score for each token using the transformer's multi-head attention layer. The difference in these scores between all the mini-batch images is calculated and provided to the Hungarian matching algorithm to generate optimal image pairs for maximum saliency. An image pair's less salient image tokens are mixed with the salient tokens. Before applying this process, TokenMixup identifies the hard images using an auxiliary classifier and skips them from the mixing procedure. Likewise to the other saliency-based image mixing methods, the augmented image labels are based on the salient information present in the output.

\subsection{Mix and Up for Object Detection}
Similar to Cutout and CutMix, MixUp is also used
to enhance the performance of object detection algorithms. One of the approaches in~\cite{mixupobj} MixUp image data to update object labels and locations. Instead of employing the basic MixUp strategy, this algorithm identifies object locations, mixes object pixels at corresponding locations in two images, and updates the object box coordinates in the output image depending on the prominent object and the image label.

\section{How does Image Mixing and Deleting improve network training?} 
\label{discussion}

\begin{figure}[tbp]
\centering
\includegraphics[width=1\columnwidth]{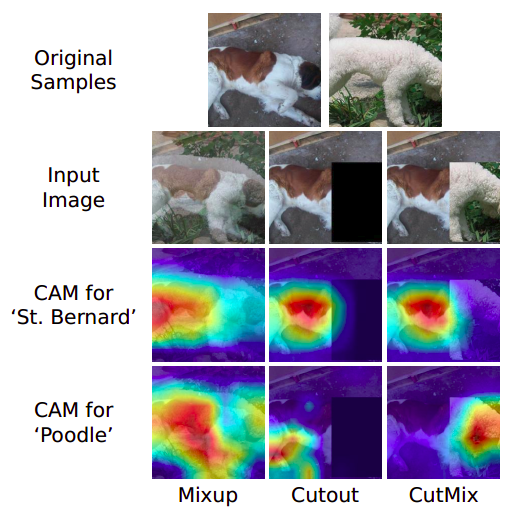}
\caption{Class Activation Maps (CAMs) for the augmented
images, which are taken from~\cite{cutmix}.}
\label{cam_fig}
\end{figure}

Understanding the underlying cause of why these techniques improve baselines’ performance is important. This builds the foundation to improve the existing approaches. Various methods are adopted to analyze standard training procedures and training with augmentations. Cutout~\cite{cutout} and CutMix~\cite{cutmix} argue that hindering image regions drives the classifier to learn from the partially visible objects and understand the overall structure. CutMix verifies this argument by showing enhanced focus toward the target class in CAMs, shown in  Figure~\ref{cam_fig}. 

In contrast, MixUp has been shown to improve the classifier’s calibration and reduce prediction uncertainty in~\cite{icapu}. To demonstrate this, the authors compared the mean of predictions against accuracy, where the confidence distribution for the MixUp trained model is evenly distributed against the standard model whose distribution is towards higher confidence, \ie, over-confidence. Similarly, the loss contours obtained for a network trained with MixUp are smooth compared to sharp contours in standard training~\cite{undermixup}. A few theoretical discussions~\cite{exp_mixup_1, exp_mixup_2} show multiple regularization effects to the standard loss with MixUp training using Taylor series approximation. MixUp introduces various regularization effects like label smoothing, Jacobian regularization, minimizing an upper bound of the adversarial loss~\cite{exp_mixup_1}, \etc Training with MixUp is equivalent to learning with structured noise in the data~\cite{exp_mixup_2}. All these theoretical and practical justifications~\cite{icapu,undermixup,exp_mixup_1,exp_mixup_2} establish better generalization and robustness against noise by MixUp training.

\vspace{2mm}
\noindent
\textbf{Strengths and Weaknesses} 
The objective of the image augmentation methods is regularization, introducing diversity in training data, and achieving robustness to occlusions and object shape changes. Augmentation methods like CutMix and MixUp add the benefit of label smoothing, whereas the additional advantage of MixUp is robustness to image noise. The benefits mentioned are common to each technique. Here, we provide the strengths and weaknesses of the augmentation methods in Tables~\ref{tab:MixDel},~\ref{tab:MixCut}~and~\ref{tab:MixUp} for each approach.

\begin{table*}[tbhp]
\caption{The strengths and weaknesses of cut and delete approaches for image classification.}
\centering
\resizebox{\textwidth}{!}{
\begin{tabular}{lll} \hline \hline
\rowcolor{gray!85} Method & Strengths & Weaknesses \\ \hline \hline
CutOut &
  \begin{tabular}[c]{@{}l@{}}Strengthen activations and increases their count, enforce \\ the network to consider diverse features in making \\ predictions, and avoid over-fitting\end{tabular} &
  \begin{tabular}[c]{@{}l@{}}Hiding regions randomly can lead to erasing \\ important regions\end{tabular} \\ \hline
Random Erasing &
  \begin{tabular}[c]{@{}l@{}}Similar benefits as cutout. Extends the experimentation \\ to object detection and shows the benefits of random \\ erasing like regularization, robustness to occlusions, etc\end{tabular} &
  \begin{tabular}[c]{@{}l@{}}Hiding regions randomly can lead to erasing \\ important regions\end{tabular} \\ \hline
Hide and Seek &
  \begin{tabular}[c]{@{}l@{}}Similar benefits as above. Erases multiple patches to\\ make the network robust to occlusions\end{tabular} &
  \begin{tabular}[c]{@{}l@{}}Severe under-fitting if too many object patches \\ are hidden\end{tabular} \\ \hline
GridMask &
  \begin{tabular}[c]{@{}l@{}}Multiple patches are deleted with even spacing between \\ them to retain a balance of deletion and retention of \\ object information\end{tabular} &
  \begin{tabular}[c]{@{}l@{}}Under-fitting if too many object patches are \\ hidden\end{tabular} \\ \hline \hline
\end{tabular}}
\label{tab:MixDel}
\end{table*}

\begin{table*}[t]
\caption{The Strengths and weaknesses  of the Cut and Mix augmentation types for image classification.}
\centering
\resizebox{\textwidth}{!}{\begin{tabular}{lll} \hline \hline
\rowcolor{gray!85} Method & Strengths & Weaknesses \\ \hline \hline
CutMix &
  \begin{tabular}[c]{@{}l@{}}Opposite to cutout, cutmix pastes a region from another \\ image along with a label value proportional to the region \\ that apparently fixes label corruption in cutout\end{tabular} &
  \begin{tabular}[c]{@{}l@{}}Label corruption is possible if an important \\ region is replaced or the background region \\ is pasted in the original image. This ends up with \\ the absence of object in the image but containing \\ the label value of the object\end{tabular} \\ \hline
Attentive CutMix &
  \begin{tabular}[c]{@{}l@{}}Only attentive regions in the second image are cutmix \\ at the same location as the original image\end{tabular} &
  \begin{tabular}[c]{@{}l@{}}Extracting attentive regions is slow as compared \\ to random selection. Attentive regions of the \\ original image are not considered during \\ replacement, this can corrupt label\end{tabular} \\ \hline
RICAP &
  \begin{tabular}[c]{@{}l@{}}Increased label smoothing effect as it creates an \\ augmented image by patching four random regions \\ from four different images with label proportion to \\ the regions\end{tabular} &
  \begin{tabular}[c]{@{}l@{}}Label corruption is possible because of \\ background information\end{tabular} \\ \hline
Mixed Example &
  \begin{tabular}[c]{@{}l@{}}Suggested 14 different linear and non-linear (vertical-\\ concat, vertical-horizontal mixup, etc) augmentations\end{tabular} &
  \begin{tabular}[c]{@{}l@{}}Only a few of the augmentation methods have \\ better performance than the baseline. No \\ performance comparison was performed with \\ the combination of multiple augmentation methods\end{tabular} \\ \hline
CowMask &
  \begin{tabular}[c]{@{}l@{}}Irregular-shaped mask similar to Friesian cow skin \\ patterns is used instead of rectangular shape. These \\ masks reduce the chances of removing consecutive \\ key regions and introduce more diversity in \\ augmented image\end{tabular} &
  \begin{tabular}[c]{@{}l@{}}Gaussian filtering is required to generate masks. \\ Blacking out too many image areas can cause \\ severe under-fitting, therefore a balance of data \\ deletion and retention is important\end{tabular} \\ \hline
ResizeMix &
  \begin{tabular}[c]{@{}l@{}}It handles label misallocation by resizing and pasting \\ the complete object on the target image randomly\end{tabular} &
  \begin{tabular}[c]{@{}l@{}}Label corruption due to background pasting of \\ source image is managed but it is possible to \\ hide key areas of the target image due to random \\ pasting\end{tabular} \\ \hline
SaliencyMix &
  \begin{tabular}[c]{@{}l@{}}The highly attentive region from the source image \\ is pasted on the target image to avoid label corruption\end{tabular} &
  \begin{tabular}[c]{@{}l@{}}Because the attentive region from the source \\ image is pasted at the corresponding location \\ in the target image, therefore it occludes the target \\ image object for some of the training data\end{tabular} \\ \hline
KeepAugment &
  \begin{tabular}[c]{@{}l@{}}This method retains salient region information of the \\ target image in augmented image\end{tabular} &
  \begin{tabular}[c]{@{}l@{}}In keepcutout and keepcutmix, source image \\ regions are deleted or pasted randomly which \\ can lead to label misallocation\end{tabular} \\ \hline
RecursiveMix &
  \begin{tabular}[c]{@{}l@{}}This method has enhanced CAMs visualization and \\learns better representations with the consistency loss\end{tabular} &
  \begin{tabular}[c]{@{}l@{}}In addition to ResizeMix weaknesses, RecursiveMix \\requires storage to maintain the history for image\\mixing and computation, i.e., RoIAlign for consistency\\loss\end{tabular} \\ \hline
LUMix &
  \begin{tabular}[c]{@{}l@{}}Reduces label misallocation in CutMix\end{tabular} &
  \begin{tabular}[c]{@{}l@{}}-\end{tabular} \\ \hline
Saliency Grafting &
  \begin{tabular}[c]{@{}l@{}}Mixes images and labels based on salient information \\with random salient patch selection to increase sample \\diversity\end{tabular} &
  \begin{tabular}[c]{@{}l@{}}Additional compute for saliency calculation\end{tabular} \\ \hline
\end{tabular}}
\label{tab:MixCut}
\end{table*}

\begin{table*}[t]
\caption{The strengths and weaknesses of image classification data augmentation approaches of Mix and UP.}
\centering

\resizebox{\textwidth}{!}{\begin{tabular}{lll} \hline \hline
\rowcolor{gray!85} Method & Strengths & Weaknesses \\ \hline \hline
MixUp &
  \begin{tabular}[c]{@{}l@{}}Additional to regularization and label smoothing, mixup\\ introduces adversarial robustness\end{tabular} &
  \begin{tabular}[c]{@{}l@{}}The advantages of mixup are unknown to image \\ segmentation. Overlapping objects in mixup \\ images can corrupt labels but not as severe as\\ cutmix\end{tabular} \\ \hline
Manifold MixUp &
  \begin{tabular}[c]{@{}l@{}}Applying mixup at hidden layers rather than at the input\\ generates tighter class representations, generalizes \\ better to novel deformations not seen during training\end{tabular} &
  \begin{tabular}[c]{@{}l@{}}Because intermediate representations are used\\ for the mixup, visualizing and explaining changes \\ in the augmented data like other augmentation \\ methods is not possible.\end{tabular} \\ \hline
AugMix &
  \begin{tabular}[c]{@{}l@{}}Preserves image semantics, creates a model that is robust\\ to corruptions and data shifts and improves calibration\\ metrics. No possibility of label misallocation\end{tabular} &
  Unknown \\ \hline
SmoothMix &
  \begin{tabular}[c]{@{}l@{}}Mixup images smoothly using a soft edge mask to avoid\\ sharp changes in pixel values\end{tabular} &
  Same as mixup \\ \hline
Co-Mixup &
  \begin{tabular}[c]{@{}l@{}}Extended mixup to multiple images, maximizes saliency\\ in the augmented image, increases smoothness in mixed \\ data\end{tabular} &
  \begin{tabular}[c]{@{}l@{}}Additional compute cost to extract saliency \\ information. The output image has broken \\ structures sometimes\end{tabular} \\ \hline
Sample Pairing &
  No additional strengths &
  \begin{tabular}[c]{@{}l@{}}Averages pixel intensities and labels of two images\\ without any diversity in generating data\end{tabular} \\ \hline
Puzzle Mix &
  \begin{tabular}[c]{@{}l@{}}Maximizes salient information in generated image,\\ creates smooth images and preserves local data statistics\end{tabular} &
  \begin{tabular}[c]{@{}l@{}}Additional compute cost to extract saliency \\ information. The output image has broken \\ structures sometimes\end{tabular} \\ \hline
SuperMix &
  \begin{tabular}[c]{@{}l@{}}Extends MixUp to multiple images, image and label \\mixing is based on salient information, and produces \\enhanced CAMs visualizations\end{tabular} &
  \begin{tabular}[c]{@{}l@{}}Optimizing MixUp masks requires backpropagating \\the teacher network multiple times\end{tabular} \\ \hline
AutoMix &
  \begin{tabular}[c]{@{}l@{}}Fixes label misallocation in MixUp, image mixing is \\based on salient information, the network training for \\classification and image mixing is end-to-end, and \\improves CAMs visualizations\end{tabular} &
  \begin{tabular}[c]{@{}l@{}}Additional compute to extract feature maps for \\saliency calculation from momentum encoder\end{tabular} \\ \hline
ReMix &
  \begin{tabular}[c]{@{}l@{}}Generates images for minority classes in imbalanced\\ datasets\end{tabular} &
  \begin{tabular}[c]{@{}l@{}}Assigns label to only minority class for the mixup\\ data\end{tabular} \\ \hline
PixMix &
  \begin{tabular}[c]{@{}l@{}}Improves model performance in various dimensions i.e.\\ robustness, consistency, calibration, adversarial, and \\ anomaly. No possibility of label misallocation\end{tabular} &
  \begin{tabular}[c]{@{}l@{}}Requires additional data containing fractals\\ or feature visualizations\end{tabular} \\ \hline
StyleMix &
  \begin{tabular}[c]{@{}l@{}}Augmented image has a mix of style and content of \\ two images.\end{tabular} &
  \begin{tabular}[c]{@{}l@{}}A style transfer network is needed to transfer style\\ of one image to the other\end{tabular} \\ \hline
AlignMixUp &
  \begin{tabular}[c]{@{}l@{}}The augmented image has pose of one image and \\ the texture of the other image. The network trained \\ with AlignMixUp has improved performance for \\ out of distribution data, calibration, and adversarial \\ robustness\end{tabular} &
  \begin{tabular}[c]{@{}l@{}}Aligning two images is a computationally \\ expensive task\end{tabular} \\ \hline
\end{tabular}}
\label{tab:MixUp}
\end{table*}

\section{Performance Comparison} 
\label{perf_comp}
We compare the augmentation approaches' performance for main-stream tasks, image classification, object detection, and fine-grained image recognition on publicly available datasets.

\begin{table*}[tbp]
\centering
\caption{Performance comparison for various image augmentation approaches for image classification problem. In the table, \enquote{R}
refers to ResNet~\cite{RNet}, \enquote{PNet} is PyramidNet~\cite{PNet}, \enquote{GNet} is GoogleNet~\cite{GNet}, and \enquote{ENet} is EfficientNet~\cite{ENet}, \enquote{FT} is fine-tuning, number of epochs are provided below the model name, where \enquote{+} sign with the number of epochs indicates the count is higher than the mentioned number, but the actual value is unknown. Similarly, \enquote{-} means the value is not known. }
\resizebox{\textwidth}{!}{
\rowcolors{2}{white}{gray!25}
\begin{tabular}{l||c|l|cc|cc|cc} \hline \hline 
\rowcolor{gray!85}&\multicolumn{1}{c}{}  &  &  \multicolumn{2}{c|}{Cifar-10} & \multicolumn{2}{c|}{Cifar-100} & \multicolumn{2}{c}{ImageNet} \\ \cline{4-9} 
   
\rowcolor{gray!85} & Code & Method &  Acc. (\%) &  (Model $\&$ epochs) &   Acc. (\%) &  (Model $\&$ epochs)  &  Acc. (\%) &  (Model $\&$ epochs) \\ \hline \hline
  
  
Cutout & $\checkmark$ & Random & 97.04 & \begin{tabular}[c]{@{}c@{}}WRN-28-10\\ 200+\end{tabular} & 81.59 & \begin{tabular}[c]{@{}c@{}}WRN-28-10\\ 200\end{tabular} & 77.1 & \begin{tabular}[c]{@{}c@{}}R-50\\ 300\end{tabular} \\ 

Random Erasing & $\checkmark$ & Random & 96.92 & \begin{tabular}[c]{@{}c@{}}WRN-28-10\\ 200+\end{tabular} & 82.27 & \begin{tabular}[c]{@{}c@{}}WRN-28-10\\ 300\end{tabular} & - & - \\ 

Hide and Seek & $\checkmark$ & Random & 96.94 & \begin{tabular}[c]{@{}c@{}}WRN-28-10\\ 200+\end{tabular} & 78.13 & \begin{tabular}[c]{@{}c@{}}R-110\\ 300\end{tabular} & 77.20 & \begin{tabular}[c]{@{}c@{}}R-50\\ 300\end{tabular} \\ 

GridMask & $\checkmark$ & Random & 97.24 & \begin{tabular}[c]{@{}c@{}}WRN-28-10\\ 200+\end{tabular} & - & - & 77.9 & \begin{tabular}[c]{@{}c@{}}R-50\\ 300\end{tabular} \\ 

CutMix & $\checkmark$ & Random & 97.10 & \begin{tabular}[c]{@{}c@{}}WRN-28-10\\ 300\end{tabular} & 83.40 & \begin{tabular}[c]{@{}c@{}}WRN-28-10\\ 300\end{tabular} & 78.6 & \begin{tabular}[c]{@{}c@{}}R-50\\ 300\end{tabular} \\ 

Attentive CutMix&  & Salient & 95.86 & \begin{tabular}[c]{@{}c@{}}ENet-B7\\ 80\end{tabular} & 78.52 & \begin{tabular}[c]{@{}c@{}}ENet-B7\\ 120\end{tabular} & - & - \\ 

RICAP & $\checkmark$ & Random & 97.18 & \begin{tabular}[c]{@{}c@{}}WRN-28-10\\ 200\end{tabular} & 82.56 & \begin{tabular}[c]{@{}c@{}}WRN-28-10\\ 200\end{tabular} & 78.62 & \begin{tabular}[c]{@{}c@{}}WRN-50-2\\ 200\end{tabular} \\ 

Mixed Example& $\checkmark$ & Random & 96.2 & \begin{tabular}[c]{@{}c@{}}R-18\\ -\end{tabular} & 80.3 & \begin{tabular}[c]{@{}c@{}}R-18\\ -\end{tabular} & - & - \\ 

CowMask & $\checkmark$ & Random & 97.44 & \begin{tabular}[c]{@{}c@{}}WRN-28-96-x2d\\ 300\end{tabular} & 84.27 & \begin{tabular}[c]{@{}c@{}}WRN-28-96-x2d\\ 300\end{tabular} & 73.94 & \begin{tabular}[c]{@{}c@{}}R-152\\ 180\end{tabular} \\ 

FMix & $\checkmark$ & Random & 96.38 & \begin{tabular}[c]{@{}c@{}}WRN-28-10\\ 200\end{tabular} & 82.03 & \begin{tabular}[c]{@{}c@{}}WRN-28-10\\ 200\end{tabular} & 77.70 & \begin{tabular}[c]{@{}c@{}}R-101\\ 90\end{tabular} \\ 

ResizeMix &  & Random & 97.60 & \begin{tabular}[c]{@{}c@{}}WRN-28-10\\ 200\end{tabular} & 84.31 & \begin{tabular}[c]{@{}c@{}}WRN-28-10\\ 200\end{tabular} & 79.00 & \begin{tabular}[c]{@{}c@{}}R-50\\ 300\end{tabular} \\ 

SaliencyMix& $\checkmark$ & Salient & 97.24 & \begin{tabular}[c]{@{}c@{}}WRN-28-10\\ 200\end{tabular} & 83.44 & \begin{tabular}[c]{@{}c@{}}WRN-28-10\\ 200\end{tabular} & 78.74 & \begin{tabular}[c]{@{}c@{}}R-50\\ 300\end{tabular} \\ 

KeepAugment &  & Salient & 97.8 & \begin{tabular}[c]{@{}c@{}}WRN-28-10\\ 300\end{tabular} & - & - & 79.1 & \begin{tabular}[c]{@{}c@{}}R-50\\ 300\end{tabular} \\ 

RecursiveMix & $\checkmark$ & Random & 97.65 & \begin{tabular}[c]{@{}c@{}}PNet-200\\ 300\end{tabular} & 81.36 & \begin{tabular}[c]{@{}c@{}}R-18\\ 200\end{tabular} & 79.2 & \begin{tabular}[c]{@{}c@{}}R-50\\ 300\end{tabular} \\ 

LUMix & & \begin{tabular}[c]{@{}l@{}}Label\\ Mixing\end{tabular} & - & - & - & - & 79.1 & \begin{tabular}[c]{@{}c@{}}R-50\\ -\end{tabular} \\ 

Saliency Grafting & & Salient & - & - & 84.68 & \begin{tabular}[c]{@{}c@{}}WRN-28-10\\ 400\end{tabular} & 77.74 & \begin{tabular}[c]{@{}c@{}}R-50\\ 100\end{tabular} \\ 

MixUp & $\checkmark$ & Random & 97.3 & \begin{tabular}[c]{@{}c@{}}WRN-28-10\\ 200\end{tabular} & 82.5 & \begin{tabular}[c]{@{}c@{}}WRN-28-10\\ 200\end{tabular} & 77.9 & \begin{tabular}[c]{@{}c@{}}R-50\\ 200\end{tabular} \\ 

Manifold-Mixup& $\checkmark$ & Random & 97.45 & \begin{tabular}[c]{@{}c@{}}WRN-28-10\\ 400\end{tabular} & 81.96 & \begin{tabular}[c]{@{}c@{}}WRN-28-10\\ 400\end{tabular} & 78.7 & \begin{tabular}[c]{@{}c@{}}R-50\\ -\end{tabular} \\ 

SmoothMix & $\checkmark$ & Random & 97.02 & \begin{tabular}[c]{@{}c@{}}PNet-200\\ 300\end{tabular} & 85.53 & \begin{tabular}[c]{@{}c@{}}PNet-200\\ 300\end{tabular} & 77.66 & \begin{tabular}[c]{@{}c@{}}R-50\\ 300\end{tabular} \\ 

Sample Pairing &  & Random & 93.07 & \begin{tabular}[c]{@{}c@{}}GNet\\ -\end{tabular} & 72.1 & \begin{tabular}[c]{@{}c@{}}GNet\\ -\end{tabular} & 70.99 & \begin{tabular}[c]{@{}c@{}}GNet\\ -\end{tabular} \\ 

AugMix & $\checkmark$ & Random & - & - & - & - & 77.6 & \begin{tabular}[c]{@{}c@{}}R-50\\ 180\end{tabular} \\ 

Co-Mixup & $\checkmark$ & Salient & - & - & 80.85 & \begin{tabular}[c]{@{}c@{}}WRN-16-8\\ 300\end{tabular} & 77.61 & \begin{tabular}[c]{@{}c@{}}R-50\\ 100\end{tabular} \\ 

Puzzle Mix & $\checkmark$ & Salient & - & - & 84.05 & \begin{tabular}[c]{@{}c@{}}WRN-28-10\\ 400\end{tabular} & 77.51 & \begin{tabular}[c]{@{}c@{}}R-50\\ -\end{tabular} \\ 

SuperMix & $\checkmark$ & Salient & - & - & 83.6 & \begin{tabular}[c]{@{}c@{}}WRN-28-10\\ -\end{tabular} & 80.8 & \begin{tabular}[c]{@{}c@{}}R-50\\ -\end{tabular} \\ 

AutoMix & $\checkmark$ & Salient & 97.34 & \begin{tabular}[c]{@{}c@{}}R-18\\ 800\end{tabular} & 85.18 & \begin{tabular}[c]{@{}c@{}}WRN-28-8\\ 800\end{tabular} & 79.25 & \begin{tabular}[c]{@{}c@{}}R-50\\ 300\end{tabular} \\ 

PixMix & $\checkmark$ & Random & - & - & - & - & 77.4 & \begin{tabular}[c]{@{}c@{}}R-50\\ 90 (FT)\end{tabular} \\ 

AlignMixUp & $\checkmark$ & \begin{tabular}[c]{@{}l@{}}Feature\\ Alignment\end{tabular} & 96.91 & \begin{tabular}[c]{@{}c@{}}WRN-16-8\\ 2000\end{tabular} & 81.23 & \begin{tabular}[c]{@{}c@{}}WRN-16-8\\ 2000\end{tabular} & 79.32 & \begin{tabular}[c]{@{}c@{}}R-50\\ 300\end{tabular} \\ 

StyleMix & $\checkmark$ & Style & 97.45 & \begin{tabular}[c]{@{}c@{}}PNet-200\\ 300\end{tabular} & 85.83 & \begin{tabular}[c]{@{}c@{}}PNet-200\\ 300\end{tabular} & 77.29 & \begin{tabular}[c]{@{}c@{}}R-50\\ 100\end{tabular} \\ \hline \hline
\end{tabular}
}
\label{img_clf_tab}
\end{table*}
\subsection{Image Classification}
The performance of image classification is evaluated on
three datasets: Cifar-10~\cite{cifar}, Cifar-100~\cite{cifar} and ImageNet~\cite{imagenet}. We collected results on these datasets from the respective articles. Results for a model common in most of the articles are selected and reported for various methodologies in Table~\ref{img_clf_tab}. The WideResnet-28-10 is largely common for the Cifar-10 and Cifar-100, while ResNet-50 is for ImageNet. We selected models with parameter count closest to the previously mentioned models for publications without reported results using these models. We can conclude about best-performing approaches for the evaluations with different models based on performance and the number of parameters. In Table~\ref{img_clf_tab}, we can observe KeepAugment~\cite{keepaugment}, Saliency Grafting~\cite{saliency_grafting}, and SuperMix~\cite{supermix} performs best for Cifar-10, Cifar-100, and ImageNet, respectively. Overall, the performance gap between salient and random mixing techniques is not very high. Hence, one can choose augmentation approaches based on the available computational resources.

\begin{table*}[!tbhp]
\caption{The improvements in accuracy for object detection using various image mixing and deleting augmentation techniques. The \enquote{R} refers to ResNet~\cite{RNet}, and \enquote{FPN} is feature pyramid network~\cite{FPN}, whereas \enquote{VGG} is from~\cite{VGG}.}
\resizebox{\textwidth}{!}{%
\begin{tabular}{lcccccl}
\hline \hline
\rowcolor{gray!85} & \begin{tabular}[c]{@{}c@{}}Backbone \\ Faster-RCNN \end{tabular} & \begin{tabular}[c]{@{}c@{}}Baseline\\ (mAP)\end{tabular} & \begin{tabular}[c]{@{}c@{}}Augmented\\ (mAP)\end{tabular} & Train Set & Test Set & \multicolumn{1}{c}{\begin{tabular}[c]{@{}c@{}}Experimentation\\ Design\end{tabular}} \\ \hline \hline
Cutout & R-50 & 76.71 & 77.17 & VOC 07+12 & VOC 07 & \begin{tabular}[c]{@{}l@{}}Backbone is trained on ImageNet using cutout\\ augmentation strategy. The Faster-RCNN backbone\\  is initialized with this newly trained model\\ and fine-tuned on train set as in~\cite{fasterrcnn}\end{tabular} \\ \hline
CutMix & R-50 & 76.71 & 78.31 & VOC 07+12 & VOC 07 & Same as above \\ \hline
MixUp & R-50 & 76.71 & 77.98 & VOC 07+12 & VOC 07 & Same as above \\ \hline
SaliencyMix & R-50 & 76.71 & 78.38 & VOC 07+12 & VOC 07 & Same as above \\ \hline
GridMask & R-50-FPN & 37.4 & 38.3 & COCO 17 & COCO 17 & \begin{tabular}[c]{@{}l@{}}Backbone trained on ImageNet is used for\\ initialization. During object detector training,\\ gridmask is applied on input after baseline\\ augmentation strategy as in~\cite{fasterrcnn}\end{tabular} \\ \hline
KeepAugment & R-50 & 38.4 & 39.5 & COCO 17 & COCO 17 & \begin{tabular}[c]{@{}l@{}}Pre-trained backbone on ImageNet with\\ keepaugment is used. The object detector\\ is fine-tuned using strategy as in~\cite{fasterrcnn}\end{tabular} \\ \hline
Random Erasing & VGG-16 & 74.8 & 76.2 & VOC 07+12 & VOC 07 & \begin{tabular}[c]{@{}l@{}}Pre-trained backbone with ImageNet weights.\\ The detector is trained with combined image\\ and object random erasing strategies keeping\\ everything similar to~\cite{fasterrcnn}\end{tabular} \\ \hline
ResizeMix & R-50 & 38.1 & 38.4 & COCO 17 & COCO 17 & Same as above \\ \hline
\end{tabular}%
}
\label{obj_det_tab}
\end{table*}

\subsection{Object Detection}
The backbones trained with these data-mixing techniques are also evaluated for object detection. The procedure involves replacing the existing backbone with a newly trained model and fine-tuning it with the object detection task, as per the standard methods mentioned in the literature. Table~\ref{obj_det_tab} shows the performance comparison for the object detection task, where the techniques are evaluated for the Faster-RCNN~\cite{fasterrcnn}. Every approach compares the standard backbone and the backbone trained with data augmentation techniques. Here, SaliencyMix outperforms for VOC-07~\cite{VOC} dataset, whereas KeepAugment is the best among all for the COCO-17~\cite{MSCOCO} dataset.

\subsection{Fine-Grained Image Recognition}
Fine-grained image classification emphasizes classifying images based on sub-class categories. For example, the class bird is further split into multiple categories based on species. More information can be seen for CUB, Cars, and Aircraft in~\cite{CUB},~\cite{cars}, and~\cite{aircraft}, respectively. Table~\ref{fine_grained_tab} shows the performance comparison for a few of the specific image augmentation techniques designed for fine-grained image recognition. All of these approaches are designed with saliency information. AttributeMix is well ahead for the CUB~\cite{CUB} and Cars~\cite{cars} dataset, whereas SnapMix succeeds for Aircraft~\cite{aircraft}.

\subsection{Vision Transformers}
We compare the image classification gains attained by image mixing in vision transformers against the baselines. The performance comparison for the ImageNet dataset is shown in Table~\ref{vision_transformers_tab}.

\begin{table}[!tbhp]
\caption{Comparison of data augmentation strategies designed 
for fine-grained image classification. The abbreviation of ResNet is \enquote{R}}
\resizebox{\columnwidth}{!}{%
\begin{tabular}{c|c|l|c|c|c}
\hline \hline
\rowcolor{gray!85} & Code & Model & CUB (\%) & Cars (\%) & Aircraft (\%) \\ \hline \hline
Attribute Mix &  & R-50 & 88.4 & 94.9 & 92.0 \\ \hline
\begin{tabular}[c]{@{}c@{}}Intra-Class \\ Part Swapping\end{tabular} &  & R-50 & 87.56 & 94.59 & 92.65 \\ \hline
SnapMix & \checkmark & R-50 & 87.75 & 94.30 & 92.80 \\ \hline 
\end{tabular}}
\label{fine_grained_tab}
\end{table}

\begin{table}[!tbhp]
\caption{Comparison of data augmentation strategies designed 
for vision transformers. The \enquote{ViT} is vision transformer~\cite{vit} and \enquote{DeiT} is data-efficient vision tranformer~\cite{deit}}
\resizebox{\columnwidth}{!}{%
\begin{tabular}{c|c|l|c|c|c}
\hline \hline
\rowcolor{gray!85} & Code & Model & Dataset & Baseline (\%) & Augmented (\%) \\ \hline \hline
TransMix & \checkmark & DeiT-S & ImageNet & 79.8 & 80.7 \\ \hline
TokenMix & \checkmark & DeiT-S & ImageNet & 79.8 & 80.8 \\ \hline
TokenMixup & \checkmark & ViT-B/16-224 & ImageNet & 81.2 & 82.32 \\ \hline 
\end{tabular}}
\label{vision_transformers_tab}
\end{table}
\section{Applications} 
\label{application}
The benefits of image mixing and deleting are not
limited to only image classification or object detection, for which they were designed initially. When used in conjunction, these algorithms have been shown to enhance the performance of other systems. A few examples are self-supervised and semi-supervised learning, unsupervised learning, adversarial training, privacy mixing, \etc The subsequent sub-sections provide a brief overview of how the incorporation of image mixing and deleting with minor tweaks have improved the performance in other applications.

\subsection{Self-Supervised Learning}
Self-supervised learning uses unlabelled data and self-generates labels for supervision. The training in self-supervised learning~\cite{mocov2,simclr,byol} generally creates multiple augmented versions of an image. This has attracted researchers to utilize image mixing and deleting methods in addition to conventional image augmentation methods. One such method, Simple Data Mixing Prior (SDMP)~\cite{SDMP}, takes a more significant image portion, resizes it, and pastes it on the other image. The augmented image is treated as a positive pair with the source image in contrastive learning and knowledge distillation. Another method, i-Mix~\cite{imix}, uses MixUp for contrastive learning to generate virtual labels. In~\cite{ssl_visualize_trans}, Manifold MixUp creates new data through feature extrapolation and interpolation. The extrapolation was only operated on positive samples to create hard positives and increase sample variance, whereas interpolation was used only for negative features to introduce sample diversity. Here, the mixing procedure ensures that the score of the hard positive samples is lower than the original positive samples.  

\subsection{Semi-Supervised Learning}
Semi-supervised learning (SSL) leverages unlabelled
data to train robust classifiers. One of the semi-supervised learning approaches, named MixMatch~\cite{SSL2}, suggested weighted MixUp, given in Eq.~\ref{stylemix_feat_vec_eq},
to ensure the augmented image is closer to the original
image. This approach is successfully employed in
other SSL techniques~\cite{remixmatch,dividemix,focalmix}. Another SSL approach~\cite{fixmatch} makes use of Cutout in association with randaugment~\cite{randaugment}, naming it Control Theory Augment (CTAugment).

\begin{equation}
\begin{gathered}
\lambda \sim \beta(\alpha, \alpha) \\
\acute{\lambda} = max(\lambda, 1 - \lambda) \\
x_{new} = \acute{\lambda}.x_a + (1-\acute{\lambda}).x_b \\
y_{new} = \acute{\lambda}.y_a + (1-\acute{\lambda}).y_b
\end{gathered}
\end{equation}

\subsection{Unsupervised Learning}
Unsupervised learning aims at learning better data representations without requiring human annotations. Recent success in self-supervised and contrastive learning~\cite{mocov2,simclr,byol} has attracted researchers to investigate image mixing and deleting for better representations in the unsupervised domain. The literature shows unsupervised algorithms using CutMix in~\cite{centerwise} and MixUp in~\cite{imix} improve the vanilla model performances simply by additional image mixing data augmentation. Another approach, Un-mix~\cite{unmix}, employs MixUp for prediction smoothness in an unsupervised setting. Instead of keeping the distance of positive pairs equal to zero, Un-mix generates a mini-batch of mixtures of positive pair augmented images. The distance between pairs is lowered and equaled to the mixing factor, either  $\lambda$ or $1 - \lambda$, based on the mixing order in a mini-batch.  

Unsupervised domain adaptation (UDA), a sub-area
of unsupervised learning utilizes MixUp to introduce
linear prediction behavior across domains. For efficient domain adaptation, from the source domain (with labeled data) to the target domain (with unlabeled data), approaches~\cite{fixbi,virtualmix,domainadaptation} mixes images from both domains. For UDA, images and labels, $x_a$ and $y_a$, are collected from the source domain, and $x_b$ and $y_b$ (pseudo label) are taken from the target domain.

\subsection{Adversarial Training}
Adversarial training trains a neural network for robustness against adversarial examples. However, while trying to achieve adversarial robustness, the neural network performance drops for the classification of clean samples. To reduce the impact of this phenomenon, some approaches~\cite{avm,iat} propose incorporating adversarial training with MixUp. Adversarial Vertex Mixup~\cite{avm} trains the network with the MixUp of clean image $x$ and adversarial vertex image $x_{av}$ generated via Eq.~\ref{vm_eq}, where perturbations $\delta_x$ are generated by Projected Gradient Descent (PGD)~\cite{pgd}. Similarly, interpolated adversarial training~\cite{iat} performs the MixUp of clean and perturbed images generated by PGD for enhanced robustness.

\begin{equation}
x_{av} = x + \gamma\delta_x
\label{vm_eq}
\end{equation}

\vspace{2mm}
\noindent
\emph{Backdoor and Targeted Attacks.} Backdoor attacks poison data by inserting a trigger patch (generally a tiny noise patch), whereas targeted attacks modify the image so that the network labels the input image as the target label. These attacks are ineffective when the model is trained with the CutMix and MixUp data augmentations as in~\cite{backdooratt,dpinstahide}.

\subsection{Privacy Preserving}
Training on edge devices produces less accurate models due to data scarcity. Edge devices also have insufficient computing resources for training and inference. To overcome these deficiencies, private data has to be transferred to the cloud. The cloud is prone to malicious activities, where attackers can exploit private data. This inspired researchers to devise solutions that preserve data privacy. Within the proposed approaches, some employ Mix $\&$ Up to achieve data privacy~\cite{instahide,datamix,xormixup}. InstaHide~\cite{instahide} mixes private images with public data k times that are later passed through random pixel sign flipping. DataMix~\cite{datamix} is a privacy-preserving inference algorithm. This algorithm mixes two private images multiple times and transmits the data to the cloud for inference. The predictions are transferred back to the edge device that performs demix operation and generates true prediction probabilities. The mixing and demixing coefficients are only known to the edge devices. XOR MixUp~\cite{xormixup} employs an iterative MixUp of private and dummy images followed by XOR for data encoding. The data is decoded on the server-side using dummy data present on the server by performing another XOR operation. After model convergence, the model is trained using decoded data whose parameters are shared with all the devices.

\subsection{Point Clouds}
Various point cloud methods~\cite{pointmixup, RSMix, pointcutmix} employ image mixing and deleting augmentations to improve performance. PointMixup~\cite{pointmixup} determines the optimal MixUp interpolation $\lambda$ value based on Earth Mover's Distance between the source pairs and the mixed point clouds. In this way, PointMixup ensures that the generated data is structurally correct instead of random noise (as in the case when MixUp is transferred directly to point clouds). Rigid Subset Mix (RSMix)~\cite{RSMix} and PointCutMix~\cite{pointcutmix} mix point clouds in CutMix style to generate augmented samples.

\subsection{Text Classification}
The applications of MixUp are also found in text
classification. Due to the discrete nature of the text,
mixing is not performed on the input, but words or
sentence embeddings~\cite{mintext,seqmix,augmentingtext}. The performance of transformers for text classification improves significantly with mixup on transformers generated embeddings~\cite{mixuptransformer}. Another approach~\cite{nonlinearmixup} suggests non-linear MixUp in the embedding space of words, where every dimension of word embedding (multi-dimensional embedding) has a different mixing parameter rather than a global value as in Eq.~\ref{mixup_eq}.

\subsection{Audio Classification}
Data mixing and deleting have been found to improve the audio classification performance~\cite{audio1,audio2,audio3}. Audio data contains various kinds of noises; it can be background noise (chirping, barking, traffic, \etc) or communication channel noise. To replicate these noises or improve audio performance generally, researchers have employed audio mixing data augmentation in~\cite{audio1, audio2} and Cut and Mix data augmentation in~\cite{audio3}.
\section{Limitations and Future Directions}
\label{limitations_and_fut_directions}
Here we discuss limitations and future directions of image mixing and deleting data augmentations:  

The speed of random image mixing and deleting data augmentations is comparable to the conventional image augmentation approaches. In contrast, these methods are slower for salient augmentations, which generally incorporate a separate network to extract salient information from input images. On the other hand, techniques like co-MixUp~\cite{comixup} and PuzzleMix~\cite{puzzlemix} train a different network to mix images based on salient information. These images are later on used for training the classification network. Training two separate networks, one for image augmentation and the other for classification, requires additional computational resources and time. Improvement to this augmentation method is designing a framework to train a single model end-to-end for augmentation and classification. 

A disadvantage of these augmentation strategies is that the generated image's structure, shape, and edges are irregular and sometimes broken. This contrasts conventional augmentation methods such as affine transformations, color variations, noise addition, \etc, where the overall structure is consistent with the input image. To avoid these drawbacks, approaches like Co-MixUp~\cite{comixup} and PuzzleMix~\cite{puzzlemix} train a model to generate a data smooth image. Although the generated image has reduced deficiencies, it still lacks smoothness and contains many structural breakages and inconsistencies around the object's edges. One future direction could be to work in this area where any number of salient objects can be included in the output image but with enhanced smoothness around the edges. 

With image mixing and deleting data augmentation, there is a possibility that the distribution of augmented images becomes different from the original training data or the target data on which the models are deployed. To avoid this problem, PointMixup~\cite{pointmixup} calculates similarities between the augmented sample and the generated sample in the point cloud domain. This can be incorporated for augmentations in the 2D image domain. 

The literature contains much less work on augmentation in feature space for images. One such work is Manifold MixUp~\cite{manifold} employing MixUp~\cite{mixup} in feature space. Analysis of using salient information in feature space, either using CutMix or MixUp, needs to be explored further in the future.  

Many efforts were devoted to understanding the improvements introduced by MixUp~\cite{icapu,undermixup,exp_mixup_1,exp_mixup_2}. The practical analysis and theoretical justifications are available in the literature proving the regularization effects of MixUp. But the literature does not contain analysis and theoretical explanations for CutOut and CutMix, other than the CAM visualizations~\cite{cutmix}. Hence, this is another future direction to explore.

A few future directions could be experimenting with corrupting foreground and background separately, rather than augmenting only background as in~\cite{keepaugment}. Approaches in MixUp corrupt both background and foreground; one can keep the foreground untouched and corrupt the background in MixUp, similar to CutMix style KeepAugment. Other than this, another future direction is to use StyleMix~\cite{stylemix} to alter the style, texture, and content in the background and foreground separately based on mixing images.
\section{Conclusion} \label{conclusion}
Data augmentation with image mixing and deleting has shown promising results, improving the accuracy of neural networks over the baselines. The improvement is not only limited to image classification but has been extended and proven for other tasks, for example, object detection, semi-supervised learning, adversarial training, \etc Whereas, seeing the performance improvements, the additional compute cost of these methods is minimal. The benefits of this type of image augmentation include label smoothing, robustness to occlusions, and adversarial robustness, in addition to regularization and increased training dataset size. This paper summarized more than 35 image mixing and deleting techniques, provided a performance comparison, and discussed their strengths $\&$ weaknesses, effects on CAMs, and the applications of these approaches. Our comparative analysis will help researchers understand each method's pros and cons, identify future research directions, and select suitable data augmentation techniques to improve the performance of their trained models. 

\bibliographystyle{spmpsci}      
\bibliography{references}   

\begin{thebibliography}{100}
\providecommand{\url}[1]{{#1}}
\providecommand{\urlprefix}{URL }
\expandafter\ifx\csname urlstyle\endcsname\relax
  \providecommand{\doi}[1]{DOI~\discretionary{}{}{}#1}\else
  \providecommand{\doi}{DOI~\discretionary{}{}{}\begingroup
  \urlstyle{rm}\Url}\fi

\bibitem{remixmatch}
Berthelot, D., Carlini, N., Cubuk, E.D., Kurakin, A., Sohn, K., Zhang, H.,
  Raffel, C.: Remixmatch: Semi-supervised learning with distribution matching
  and augmentation anchoring.
\newblock In: International Conference on Learning Representations (2020)

\bibitem{SSL2}
Berthelot, D., Carlini, N., Goodfellow, I., Papernot, N., Oliver, A., Raffel,
  C.A.: Mixmatch: A holistic approach to semi-supervised learning.
\newblock Advances in neural information processing systems \textbf{32} (2019)

\bibitem{backdooratt}
Borgnia, E., Cherepanova, V., Fowl, L., Ghiasi, A., Geiping, J., Goldblum, M.,
  Goldstein, T., Gupta, A.: Strong data augmentation sanitizes poisoning and
  backdoor attacks without an accuracy tradeoff.
\newblock In: {ICASSP}, pp. 3855--3859. {IEEE} (2021)

\bibitem{dpinstahide}
Borgnia, E., Geiping, J., Cherepanova, V., Fowl, L., Gupta, A., Ghiasi, A.,
  Huang, F., Goldblum, M., Goldstein, T.: Dp-instahide: Provably defusing
  poisoning and backdoor attacks with differentially private data
  augmentations.
\newblock arXiv preprint arXiv:2103.02079  (2021)

\bibitem{mixupobj}
Bouabid, S., Delaitre, V.: Mixup regularization for region proposal based
  object detectors.
\newblock arXiv preprint arXiv:2003.02065  (2020)

\bibitem{exp_mixup_2}
Carratino, L., Ciss{\'e}, M., Jenatton, R., Vert, J.P.: On mixup
  regularization.
\newblock arXiv preprint arXiv:2006.06049  (2020)

\bibitem{mintext}
Chen, J., Yang, Z., Yang, D.: Mixtext: Linguistically-informed interpolation of
  hidden space for semi-supervised text classification.
\newblock In: {ACL}, pp. 2147--2157. Association for Computational Linguistics
  (2020)

\bibitem{transmix}
Chen, J.N., Sun, S., He, J., Torr, P.H., Yuille, A., Bai, S.: Transmix: Attend
  to mix for vision transformers.
\newblock In: Proceedings of the IEEE/CVF Conference on Computer Vision and
  Pattern Recognition, pp. 12135--12144 (2022)

\bibitem{deeplab}
Chen, L.C., Papandreou, G., Kokkinos, I., Murphy, K., Yuille, A.L.: Deeplab:
  Semantic image segmentation with deep convolutional nets, atrous convolution,
  and fully connected crfs.
\newblock IEEE transactions on pattern analysis and machine intelligence
  \textbf{40}(4), 834--848 (2017)

\bibitem{gridmask}
Chen, P., Liu, S., Zhao, H., Jia, J.: Gridmask data augmentation.
\newblock arXiv preprint arXiv:2001.04086  (2020)

\bibitem{simclr}
Chen, T., Kornblith, S., Swersky, K., Norouzi, M., Hinton, G.E.: Big
  self-supervised models are strong semi-supervised learners.
\newblock In: NeurIPS (2020)

\bibitem{mocov2}
Chen, X., Fan, H., Girshick, R., He, K.: Improved baselines with momentum
  contrastive learning.
\newblock Technical Report  (2020)

\bibitem{pointmixup}
Chen, Y., Hu, V.T., Gavves, E., Mensink, T., Mettes, P., Yang, P., Snoek, C.G.:
  Pointmixup: Augmentation for point clouds.
\newblock In: European Conference on Computer Vision, pp. 330--345. Springer
  (2020)

\bibitem{tokenmixup}
Choi, H.K., Choi, J., Kim, H.J.: Tokenmixup: Efficient attention-guided
  token-level data augmentation for transformers.
\newblock In: Advances in Neural Information Processing Systems (2022)

\bibitem{remix}
Chou, H.P., Chang, S.C., Pan, J.Y., Wei, W., Juan, D.C.: Remix: Rebalanced
  mixup.
\newblock In: European Conference on Computer Vision, pp. 95--110. Springer
  (2020)

\bibitem{autoaugment}
Cubuk, E.D., Zoph, B., Mane, D., Vasudevan, V., Le, Q.V.: Autoaugment: Learning
  augmentation strategies from data.
\newblock In: Proceedings of the IEEE/CVF Conference on Computer Vision and
  Pattern Recognition, pp. 113--123 (2019)

\bibitem{randaugment}
Cubuk, E.D., Zoph, B., Shlens, J., Le, Q.V.: Randaugment: Practical automated
  data augmentation with a reduced search space.
\newblock In: Proceedings of the IEEE/CVF Conference on Computer Vision and
  Pattern Recognition Workshops, pp. 702--703 (2020)

\bibitem{sinkhorn}
Cuturi, M.: Sinkhorn distances: Lightspeed computation of optimal transport.
\newblock Advances in neural information processing systems \textbf{26} (2013)

\bibitem{supermix}
Dabouei, A., Soleymani, S., Taherkhani, F., Nasrabadi, N.M.: Supermix:
  Supervising the mixing data augmentation.
\newblock In: Proceedings of the IEEE/CVF Conference on Computer Vision and
  Pattern Recognition, pp. 13794--13803 (2021)

\bibitem{cutout}
DeVries, T., Taylor, G.W.: Improved regularization of convolutional neural
  networks with cutout.
\newblock arXiv preprint arXiv:1708.04552  (2017)

\bibitem{vits}
Dosovitskiy, A., Beyer, L., Kolesnikov, A., Weissenborn, D., Zhai, X.,
  Unterthiner, T., Dehghani, M., Minderer, M., Heigold, G., Gelly, S.,
  Uszkoreit, J., Houlsby, N.: An image is worth 16x16 words: Transformers for
  image recognition at scale.
\newblock In: {ICLR}. OpenReview.net (2021)

\bibitem{vit}
Dosovitskiy, A., Beyer, L., Kolesnikov, A., Weissenborn, D., Zhai, X.,
  Unterthiner, T., Dehghani, M., Minderer, M., Heigold, G., Gelly, S.,
  Uszkoreit, J., Houlsby, N.: An image is worth 16x16 words: Transformers for
  image recognition at scale.
\newblock In: ICLR (2021)

\bibitem{VCAugment1}
Dvornik, N., Mairal, J., Schmid, C.: Modeling visual context is key to
  augmenting object detection datasets.
\newblock In: Proceedings of the European Conference on Computer Vision (ECCV),
  pp. 364--380 (2018)

\bibitem{cutpasteandlearn}
Dwibedi, D., Misra, I., Hebert, M.: Cut, paste and learn: Surprisingly easy
  synthesis for instance detection.
\newblock In: Proceedings of the IEEE International Conference on Computer
  Vision, pp. 1301--1310 (2017)

\bibitem{VOC}
Everingham, M., Van~Gool, L., Williams, C.K., Winn, J., Zisserman, A.: The
  pascal visual object classes (voc) challenge.
\newblock International journal of computer vision \textbf{88}(2), 303--338
  (2010)

\bibitem{cowmask}
French, G., Oliver, A., Salimans, T.: Milking cowmask for semi-supervised image
  classification.
\newblock In: {VISIGRAPP} {(5:} {VISAPP)}, pp. 75--84. {SCITEPRESS} (2022)

\bibitem{keepaugment}
Gong, C., Wang, D., Li, M., Chandra, V., Liu, Q.: Keepaugment: A simple
  information-preserving data augmentation approach.
\newblock In: Proceedings of the IEEE/CVF conference on computer vision and
  pattern recognition, pp. 1055--1064 (2021)

\bibitem{GANs}
Goodfellow, I., Pouget-Abadie, J., Mirza, M., Xu, B., Warde-Farley, D., Ozair,
  S., Courville, A., Bengio, Y.: Generative adversarial networks.
\newblock Communications of the ACM \textbf{63}(11), 139--144 (2020)

\bibitem{byol}
Grill, J., Strub, F., Altch{\'{e}}, F., Tallec, C., Richemond, P.H.,
  Buchatskaya, E., Doersch, C., Pires, B.{\'{A}}., Guo, Z., Azar, M.G., Piot,
  B., Kavukcuoglu, K., Munos, R., Valko, M.: Bootstrap your own latent - {A}
  new approach to self-supervised learning.
\newblock In: NeurIPS (2020)

\bibitem{nonlinearmixup}
Guo, H.: Nonlinear mixup: Out-of-manifold data augmentation for text
  classification.
\newblock In: {AAAI}, pp. 4044--4051. {AAAI} Press (2020)

\bibitem{augmentingtext}
Guo, H., Mao, Y., Zhang, R.: Augmenting data with mixup for sentence
  classification: An empirical study.
\newblock arXiv preprint arXiv:1905.08941  (2019)

\bibitem{PNet}
Han, D., Kim, J., Kim, J.: Deep pyramidal residual networks.
\newblock In: Proceedings of the IEEE conference on computer vision and pattern
  recognition, pp. 5927--5935 (2017)

\bibitem{fmix}
Harris, E., Marcu, A., Painter, M., Niranjan, M., Pr{\"u}gel-Bennett, A., Hare,
  J.: Fmix: Enhancing mixed sample data augmentation.
\newblock arXiv preprint arXiv:2002.12047  (2020)

\bibitem{mask_rcnn}
He, K., Gkioxari, G., Doll{\'a}r, P., Girshick, R.: Mask r-cnn.
\newblock In: Proceedings of the IEEE international conference on computer
  vision, pp. 2961--2969 (2017)

\bibitem{RNet}
He, K., Zhang, X., Ren, S., Sun, J.: Deep residual learning for image
  recognition.
\newblock In: Proceedings of the IEEE conference on computer vision and pattern
  recognition, pp. 770--778 (2016)

\bibitem{augmix}
Hendrycks, D., Mu, N., Cubuk, E.D., Zoph, B., Gilmer, J., Lakshminarayanan, B.:
  Augmix: {A} simple data processing method to improve robustness and
  uncertainty.
\newblock In: {ICLR}. OpenReview.net (2020)

\bibitem{pixmix}
Hendrycks, D., Zou, A., Mazeika, M., Tang, L., Li, B., Song, D., Steinhardt,
  J.: Pixmix: Dreamlike pictures comprehensively improve safety measures.
\newblock In: Proceedings of the IEEE/CVF Conference on Computer Vision and
  Pattern Recognition, pp. 16783--16792 (2022)

\bibitem{stylemix}
Hong, M., Choi, J., Kim, G.: Stylemix: Separating content and style for
  enhanced data augmentation.
\newblock In: Proceedings of the IEEE/CVF Conference on Computer Vision and
  Pattern Recognition, pp. 14862--14870 (2021)

\bibitem{snapmix}
Huang, S., Wang, X., Tao, D.: Snapmix: Semantically proportional mixing for
  augmenting fine-grained data.
\newblock In: Proceedings of the AAAI Conference on Artificial Intelligence,
  vol.~35, pp. 1628--1636 (2021)

\bibitem{adain}
Huang, X., Belongie, S.: Arbitrary style transfer in real-time with adaptive
  instance normalization.
\newblock In: Proceedings of the IEEE international conference on computer
  vision, pp. 1501--1510 (2017)

\bibitem{instahide}
Huang, Y., Song, Z., Li, K., Arora, S.: Instahide: Instance-hiding schemes for
  private distributed learning.
\newblock In: {ICML}, \emph{Proceedings of Machine Learning Research}, vol.
  119, pp. 4507--4518. {PMLR} (2020)

\bibitem{samplepairing}
Inoue, H.: Data augmentation by pairing samples for images classification.
\newblock arXiv preprint arXiv:1801.02929  (2018)

\bibitem{surveyimaug2}
Khalifa, N.E., Loey, M., Mirjalili, S.: A comprehensive survey of recent trends
  in deep learning for digital images augmentation.
\newblock Artificial Intelligence Review pp. 1--27 (2021)

\bibitem{audio3}
Kim, G., Han, D.K., Ko, H.: Specmix: A mixed sample data augmentation method
  for training with time-frequency domain features.
\newblock arXiv preprint arXiv:2108.03020  (2021)

\bibitem{comixup}
Kim, J., Choo, W., Jeong, H., Song, H.O.: Co-mixup: Saliency guided joint mixup
  with supermodular diversity.
\newblock In: {ICLR}. OpenReview.net (2021)

\bibitem{puzzlemix}
Kim, J.H., Choo, W., Song, H.O.: Puzzle mix: Exploiting saliency and local
  statistics for optimal mixup.
\newblock In: International Conference on Machine Learning, pp. 5275--5285.
  PMLR (2020)

\bibitem{cars}
Krause, J., Stark, M., Deng, J., Fei-Fei, L.: 3d object representations for
  fine-grained categorization.
\newblock In: Proceedings of the IEEE international conference on computer
  vision workshops, pp. 554--561 (2013)

\bibitem{cifar}
Krizhevsky, A., Hinton, G., et~al.: Learning multiple layers of features from
  tiny images.
\newblock Technical Report  (2009)

\bibitem{clf}
Krizhevsky, A., Sutskever, I., Hinton, G.E.: Imagenet classification with deep
  convolutional neural networks.
\newblock Advances in neural information processing systems \textbf{25},
  1097--1105 (2012)

\bibitem{hidenseek}
Kumar~Singh, K., Jae~Lee, Y.: Hide-and-seek: Forcing a network to be meticulous
  for weakly-supervised object and action localization.
\newblock In: Proceedings of the IEEE International Conference on Computer
  Vision, pp. 3524--3533 (2017)

\bibitem{iat}
Lamb, A., Verma, V., Kannala, J., Bengio, Y.: Interpolated adversarial
  training: Achieving robust neural networks without sacrificing too much
  accuracy.
\newblock In: AISec@CCS, pp. 95--103. {ACM} (2019)

\bibitem{RSMix}
Lee, D., Lee, J., Lee, J., Lee, H., Lee, M., Woo, S., Lee, S.: Regularization
  strategy for point cloud via rigidly mixed sample.
\newblock In: Proceedings of the IEEE/CVF Conference on Computer Vision and
  Pattern Recognition, pp. 15900--15909 (2021)

\bibitem{smoothmix}
Lee, J.H., Zaigham~Zaheer, M., Astrid, M., Lee, S.I.: Smoothmix: A simple yet
  effective data augmentation to train robust classifiers.
\newblock In: Proceedings of the IEEE/CVF Conference on Computer Vision and
  Pattern Recognition Workshops, pp. 756--757 (2020)

\bibitem{imix}
Lee, K., Zhu, Y., Sohn, K., Li, C., Shin, J., Lee, H.: i-mix: {A}
  domain-agnostic strategy for contrastive representation learning.
\newblock In: {ICLR}. OpenReview.net (2021)

\bibitem{avm}
Lee, S., Lee, H., Yoon, S.: Adversarial vertex mixup: Toward better
  adversarially robust generalization.
\newblock In: {CVPR}, pp. 269--278. Computer Vision Foundation / {IEEE} (2020)

\bibitem{regularization}
Lever, J., Krzywinski, M., Altman, N.: Regularization (2016)

\bibitem{overfitting2}
Li, H., Li, J., Guan, X., Liang, B., Lai, Y., Luo, X.: Research on overfitting
  of deep learning.
\newblock In: 2019 15th International Conference on Computational Intelligence
  and Security (CIS), pp. 78--81. IEEE (2019)

\bibitem{attributemix}
Li, H., Zhang, X., Tian, Q., Xiong, H.: Attribute mix: semantic data
  augmentation for fine grained recognition.
\newblock In: 2020 IEEE International Conference on Visual Communications and
  Image Processing (VCIP), pp. 243--246. IEEE (2020)

\bibitem{centerwise}
Li, H., Zhang, X., Xiong, H.: Center-wise local image mixture for contrastive
  representation learning.
\newblock In: {BMVC}, p. 369. {BMVA} Press (2021)

\bibitem{dividemix}
Li, J., Socher, R., Hoi, S.C.H.: Dividemix: Learning with noisy labels as
  semi-supervised learning.
\newblock In: {ICLR}. OpenReview.net (2020)

\bibitem{undermixup}
Liang, D., Yang, F., Zhang, T., Yang, P.: Understanding mixup training methods.
\newblock {IEEE} Access \textbf{6}, 58774--58783 (2018)

\bibitem{fastaugment}
Lim, S., Kim, I., Kim, T., Kim, C., Kim, S.: Fast autoaugment.
\newblock Advances in Neural Information Processing Systems \textbf{32} (2019)

\bibitem{noisyfeaturemixup}
Lim, S.H., Erichson, N.B., Utrera, F., Xu, W., Mahoney, M.W.: Noisy feature
  mixup.
\newblock In: International Conference on Learning Representations (2022)

\bibitem{FPN}
Lin, T.Y., Doll{\'a}r, P., Girshick, R., He, K., Hariharan, B., Belongie, S.:
  Feature pyramid networks for object detection.
\newblock In: Proceedings of the IEEE conference on computer vision and pattern
  recognition, pp. 2117--2125 (2017)

\bibitem{MSCOCO}
Lin, T.Y., Maire, M., Belongie, S., Hays, J., Perona, P., Ramanan, D.,
  Doll{\'a}r, P., Zitnick, C.L.: Microsoft coco: Common objects in context.
\newblock In: European conference on computer vision, pp. 740--755. Springer
  (2014)

\bibitem{tokenmix}
Liu, J., Liu, B., Zhou, H., Li, H., Liu, Y.: Tokenmix: Rethinking image mixing
  for data augmentation in vision transformers.
\newblock In: European Conference on Computer Vision, pp. 455--471. Springer
  (2022)

\bibitem{automix}
Liu, Z., Li, S., Wu, D., Liu, Z., Chen, Z., Wu, L., Li, S.Z.: Automix:
  Unveiling the power of mixup for stronger classifiers.
\newblock In: European Conference on Computer Vision, pp. 441--458. Springer
  (2022)

\bibitem{datamix}
Liu, Z., Wu, Z., Gan, C., Zhu, L., Han, S.: Datamix: Efficient
  privacy-preserving edge-cloud inference.
\newblock In: {ECCV} {(11)}, \emph{Lecture Notes in Computer Science}, vol.
  12356, pp. 578--595. Springer (2020)

\bibitem{pgd}
Madry, A., Makelov, A., Schmidt, L., Tsipras, D., Vladu, A.: Towards deep
  learning models resistant to adversarial attacks.
\newblock In: {ICLR} (Poster). OpenReview.net (2018)

\bibitem{aircraft}
Maji, S., Rahtu, E., Kannala, J., Blaschko, M., Vedaldi, A.: Fine-grained
  visual classification of aircraft.
\newblock arXiv preprint arXiv:1306.5151  (2013)

\bibitem{virtualmix}
Mao, X., Ma, Y., Yang, Z., Chen, Y., Li, Q.: Virtual mixup training for
  unsupervised domain adaptation.
\newblock arXiv preprint arXiv:1905.04215  (2019)

\bibitem{fixbi}
Na, J., Jung, H., Chang, H.J., Hwang, W.: Fixbi: Bridging domain spaces for
  unsupervised domain adaptation.
\newblock In: Proceedings of the IEEE/CVF Conference on Computer Vision and
  Pattern Recognition, pp. 1094--1103 (2021)

\bibitem{overfitting}
Naveed, H., Jafri, F., Javed, K., Babri, H.A.: Driver activity recognition by
  learning spatiotemporal features of pose and human object interaction.
\newblock Journal of Visual Communication and Image Representation \textbf{77},
  103135 (2021)

\bibitem{saliency_grafting}
Park, J., Yang, J.Y., Shin, J., Hwang, S.J., Yang, E.: Saliency grafting:
  Innocuous attribution-guided mixup with calibrated label mixing.
\newblock In: Proceedings of the AAAI Conference on Artificial Intelligence,
  vol.~36, pp. 7957--7965 (2022)

\bibitem{regmixup}
Pinto, F., Yang, H., Lim, S.N., Torr, P., Dokania, P.K.: Using mixup as a
  regularizer can surprisingly improve accuracy \& out-of-distribution
  robustness.
\newblock In: Advances in Neural Information Processing Systems (2022)

\bibitem{resizemix}
Qin, J., Fang, J., Zhang, Q., Liu, W., Wang, X., Wang, X.: Resizemix: Mixing
  data with preserved object information and true labels.
\newblock arXiv preprint arXiv:2012.11101  (2020)

\bibitem{yolov3}
Redmon, J., Farhadi, A.: Yolov3: An incremental improvement.
\newblock Technical Report  (2018)

\bibitem{fasterrcnn}
Ren, S., He, K., Girshick, R.B., Sun, J.: Faster {R-CNN:} towards real-time
  object detection with region proposal networks.
\newblock {IEEE} Trans. Pattern Anal. Mach. Intell. \textbf{39}(6), 1137--1149
  (2017)

\bibitem{SDMP}
Ren, S., Wang, H., Gao, Z., He, S., Yuille, A., Zhou, Y., Xie, C.: A simple
  data mixing prior for improving self-supervised learning.
\newblock In: Proceedings of the IEEE/CVF Conference on Computer Vision and
  Pattern Recognition, pp. 14595--14604 (2022)

\bibitem{imagenet}
Russakovsky, O., Deng, J., Su, H., Krause, J., Satheesh, S., Ma, S., Huang, Z.,
  Karpathy, A., Khosla, A., Bernstein, M., et~al.: Imagenet large scale visual
  recognition challenge.
\newblock International journal of computer vision \textbf{115}(3), 211--252
  (2015)

\bibitem{unmix}
Shen, Z., Liu, Z., Liu, Z., Savvides, M., Darrell, T., Xing, E.: Un-mix:
  Rethinking image mixtures for unsupervised visual representation learning.
\newblock In: Proceedings of the AAAI Conference on Artificial Intelligence,
  vol.~36, pp. 2216--2224 (2022)

\bibitem{xormixup}
Shin, M., Hwang, C., Kim, J., Park, J., Bennis, M., Kim, S.L.: Xor mixup:
  Privacy-preserving data augmentation for one-shot federated learning.
\newblock arXiv preprint arXiv:2006.05148  (2020)

\bibitem{surveyimaug}
Shorten, C., Khoshgoftaar, T.M.: A survey on image data augmentation for deep
  learning.
\newblock Journal of Big Data \textbf{6}(1), 1--48 (2019)

\bibitem{VGG}
Simonyan, K., Zisserman, A.: Very deep convolutional networks for large-scale
  image recognition.
\newblock In: {ICLR} (2015)

\bibitem{fixmatch}
Sohn, K., Berthelot, D., Carlini, N., Zhang, Z., Zhang, H., Raffel, C., Cubuk,
  E.D., Kurakin, A., Li, C.: Fixmatch: Simplifying semi-supervised learning
  with consistency and confidence.
\newblock In: NeurIPS (2020)

\bibitem{dropout}
Srivastava, N., Hinton, G., Krizhevsky, A., Sutskever, I., Salakhutdinov, R.:
  Dropout: a simple way to prevent neural networks from overfitting.
\newblock The journal of machine learning research \textbf{15}(1), 1929--1958
  (2014)

\bibitem{mixedexample}
Summers, C., Dinneen, M.J.: Improved mixed-example data augmentation.
\newblock In: 2019 IEEE Winter Conference on Applications of Computer Vision
  (WACV), pp. 1262--1270. IEEE (2019)

\bibitem{mixuptransformer}
Sun, L., Xia, C., Yin, W., Liang, T., Yu, P.S., He, L.: Mixup-transformer:
  Dynamic data augmentation for {NLP} tasks.
\newblock In: {COLING}, pp. 3436--3440. International Committee on
  Computational Linguistics (2020)

\bibitem{lumix}
Sun, S., Chen, J.N., He, R., Yuille, A., Torr, P., Bai, S.: Lumix: Improving
  mixup by better modelling label uncertainty.
\newblock arXiv preprint arXiv:2211.15846  (2022)

\bibitem{GNet}
Szegedy, C., Liu, W., Jia, Y., Sermanet, P., Reed, S., Anguelov, D., Erhan, D.,
  Vanhoucke, V., Rabinovich, A.: Going deeper with convolutions.
\newblock In: Proceedings of the IEEE conference on computer vision and pattern
  recognition, pp. 1--9 (2015)

\bibitem{ricap}
Takahashi, R., Matsubara, T., Uehara, K.: Ricap: Random image cropping and
  patching data augmentation for deep cnns.
\newblock In: Asian Conference on Machine Learning, pp. 786--798 (2018)

\bibitem{ENet}
Tan, M., Le, Q.: Efficientnet: Rethinking model scaling for convolutional
  neural networks.
\newblock In: International Conference on Machine Learning, pp. 6105--6114.
  PMLR (2019)

\bibitem{icapu}
Thulasidasan, S., Chennupati, G., Bilmes, J.A., Bhattacharya, T., Michalak, S.:
  On mixup training: Improved calibration and predictive uncertainty for deep
  neural networks.
\newblock In: NeurIPS, pp. 13888--13899 (2019)

\bibitem{deit}
Touvron, H., Cord, M., Douze, M., Massa, F., Sablayrolles, A., J{\'e}gou, H.:
  Training data-efficient image transformers \& distillation through attention.
\newblock In: International Conference on Machine Learning, pp. 10347--10357.
  PMLR (2021)

\bibitem{saliencymix}
Uddin, A.F.M.S., Monira, M.S., Shin, W., Chung, T., Bae, S.: Saliencymix: {A}
  saliency guided data augmentation strategy for better regularization.
\newblock In: {ICLR}. OpenReview.net (2021)

\bibitem{alignmixup}
Venkataramanan, S., Kijak, E., Amsaleg, L., Avrithis, Y.: Alignmixup: Improving
  representations by interpolating aligned features.
\newblock In: Proceedings of the IEEE/CVF Conference on Computer Vision and
  Pattern Recognition, pp. 19174--19183 (2022)

\bibitem{manifold}
Verma, V., Lamb, A., Beckham, C., Najafi, A., Mitliagkas, I., Lopez-Paz, D.,
  Bengio, Y.: Manifold mixup: Better representations by interpolating hidden
  states.
\newblock In: International Conference on Machine Learning, pp. 6438--6447.
  PMLR (2019)

\bibitem{attentivecutmix}
Walawalkar, D., Shen, Z., Liu, Z., Savvides, M.: Attentive cutmix: An enhanced
  data augmentation approach for deep learning based image classification.
\newblock In: ICASSP 2020-2020 IEEE International Conference on Acoustics,
  Speech and Signal Processing (ICASSP), pp. 3642--3646. IEEE (2020)

\bibitem{focalmix}
Wang, D., Zhang, Y., Zhang, K., Wang, L.: Focalmix: Semi-supervised learning
  for 3d medical image detection.
\newblock In: Proceedings of the IEEE/CVF Conference on Computer Vision and
  Pattern Recognition, pp. 3951--3960 (2020)

\bibitem{ASDN}
Wang, X., Shrivastava, A., Gupta, A.: A-fast-rcnn: Hard positive generation via
  adversary for object detection.
\newblock In: Proceedings of the IEEE conference on computer vision and pattern
  recognition, pp. 2606--2615 (2017)

\bibitem{CUB}
Welinder, P., Branson, S., Mita, T., Wah, C., Schroff, F., Belongie, S.,
  Perona, P.: Caltech-ucsd birds 200.
\newblock Technical Report  (2010)

\bibitem{domainadaptation}
Yan, S., Song, H., Li, N., Zou, L., Ren, L.: Improve unsupervised domain
  adaptation with mixup training.
\newblock arXiv preprint arXiv:2001.00677  (2020)

\bibitem{recursivemix}
Yang, L., Li, X., Zhao, B., Song, R., Yang, J.: Recursivemix: Mixed learning
  with history.
\newblock In: Advances in Neural Information Processing Systems (2022)

\bibitem{overfitting3}
Ying, X.: An overview of overfitting and its solutions.
\newblock In: Journal of Physics: Conference Series, vol. 1168, p. 022022. IOP
  Publishing (2019)

\bibitem{cutmix}
Yun, S., Han, D., Oh, S.J., Chun, S., Choe, J., Yoo, Y.: Cutmix: Regularization
  strategy to train strong classifiers with localizable features.
\newblock In: Proceedings of the IEEE International Conference on Computer
  Vision, pp. 6023--6032 (2019)

\bibitem{mixup}
Zhang, H., Ciss{\'{e}}, M., Dauphin, Y.N., Lopez{-}Paz, D.: mixup: Beyond
  empirical risk minimization.
\newblock In: {ICLR} (Poster). OpenReview.net (2018)

\bibitem{pointcutmix}
Zhang, J., Chen, L., Ouyang, B., Liu, B., Zhu, J., Chen, Y., Meng, Y., Wu, D.:
  Pointcutmix: Regularization strategy for point cloud classification.
\newblock Neurocomputing \textbf{505}, 58--67 (2022)

\bibitem{exp_mixup_1}
Zhang, L., Deng, Z., Kawaguchi, K., Ghorbani, A., Zou, J.: How does mixup help
  with robustness and generalization?
\newblock In: {ICLR}. OpenReview.net (2021)

\bibitem{intraPartSwap}
Zhang, L., Huang, S., Liu, W.: Intra-class part swapping for fine-grained image
  classification.
\newblock In: Proceedings of the IEEE/CVF Winter Conference on Applications of
  Computer Vision, pp. 3209--3218 (2021)

\bibitem{seqmix}
Zhang, R., Yu, Y., Zhang, C.: Seqmix: Augmenting active sequence labeling via
  sequence mixup.
\newblock Proceedings of the Conference on Empirical Methods in Natural
  Language Processing (EMNLP)  (2020)

\bibitem{semanticseg}
Zhang, Y., Qiu, Z., Yao, T., Liu, D., Mei, T.: Fully convolutional adaptation
  networks for semantic segmentation.
\newblock In: Proceedings of the IEEE Conference on Computer Vision and Pattern
  Recognition, pp. 6810--6818 (2018)

\bibitem{audio1}
Zhang, Z., Xu, S., Cao, S., Zhang, S.: Deep convolutional neural network with
  mixup for environmental sound classification.
\newblock In: Chinese conference on pattern recognition and computer vision
  (prcv), pp. 356--367. Springer (2018)

\bibitem{rerase}
Zhong, Z., Zheng, L., Kang, G., Li, S., Yang, Y.: Random erasing data
  augmentation.
\newblock In: AAAI, pp. 13001--13008 (2020)

\bibitem{ssl_visualize_trans}
Zhu, R., Zhao, B., Liu, J., Sun, Z., Chen, C.W.: Improving contrastive learning
  by visualizing feature transformation.
\newblock In: Proceedings of the IEEE/CVF International Conference on Computer
  Vision, pp. 10306--10315 (2021)

\bibitem{audio2}
Zhu, Y., Ko, T., Mak, B.: Mixup learning strategies for text-independent
  speaker verification.
\newblock In: Interspeech, pp. 4345--4349 (2019)

\end{thebibliography}
\end{document}